\documentclass[Afour,sageh,times]{sagej}
\usepackage[linesnumbered,ruled,vlined]{algorithm2e}
\usepackage{multirow,makecell}
\usepackage[colorlinks,bookmarksopen,bookmarksnumbered,citecolor=red,urlcolor=red]{hyperref}

\allowdisplaybreaks
\makeatletter
\renewcommand{\@maketitle}{%
  \begingroup
  \sffamily\raggedright
  {\titlesize\bfseries\@title\par}
  \vspace{0.8em}
  {\large\bfseries\@author\par}
  \vspace{1.4em}
  \noindent\usebox\absbox\par
  \vspace{1em}
  \noindent\@keywords\par
  \endgroup
  \vspace{1.6em}}
\let\ps@title\ps@plain
\makeatother
\begin{document}

\title{SDF-BA: Geometric Bundle Adjustment in Signed Distance Fields for LiDAR SLAM}

\author{Yingyu Wang\affilnum{1}, Tiancheng Li\affilnum{2}, Lan Wu\affilnum{3}, Liang Zhao\affilnum{1}, and Shoudong Huang\affilnum{2}}

\affiliation{\affilnum{1}School of Informatics, The University of Edinburgh, UK \\
\affilnum{2}Robotics Institute, University of Technology Sydney, Australia\\
\affilnum{3}School of Engineering, The University of Western Australia, Australia
}

\corrauth{
Shoudong Huang, Robotics Institute, University of Technology Sydney, 15 Broadway, Ultimo, NSW 2007, Australia; and Liang Zhao, School of Informatics, The University of Edinburgh, Edinburgh EH8 9BT, UK.}

\email{Shoudong.Huang@uts.edu.au; Liang.Zhao@ed.ac.uk}

\begin{abstract}

Signed distance fields (SDFs) provide continuous representations for dense geometric mapping, but in conventional tracking and fusion pipelines, pose errors can become embedded in the reconstructed geometry when observations are integrated using imperfect pose estimates. We present SDF-BA, a geometric bundle adjustment formulation that jointly optimizes robot poses and truncated signed distance field (TSDF) values from LiDAR range observations. A central formulation challenge is to establish geometric observation--map constraints from range observations, since signed distances inferred from locally observed surfaces do not, in general, coincide with the corresponding values of the global SDF. To address this challenge, we exploit the geometric relationships among signed distances, field gradients, and surface locations to construct pose-dependent observation--map constraints directly in the SDF domain. The resulting nonlinear least squares formulation uses range
observations to jointly constrain the robot trajectory, the
zero-level surface, and off-surface TSDF values within a unified geometric objective, rather than treating the map solely as a fixed reference for pose estimation. To maintain computational tractability, we introduce a compact
TSDF optimization state. For large-scale environments, we integrate SDF-BA into a submap-based SLAM framework that jointly optimizes robot poses and TSDF values within local submaps and aligns the submaps through direct SDF-based graph optimization. Extensive experiments on public LiDAR datasets evaluate trajectory accuracy, reconstruction quality, robustness, and computational efficiency. The results show improved trajectory accuracy and reconstruction quality over representative state-of-the-art approaches, while ablation studies demonstrate the benefits of joint pose--map optimization and geometric constraints beyond the zero-level surface. The source code will be made publicly available upon paper acceptance.

\end{abstract}

\keywords{Bundle adjustment, LiDAR SLAM, signed distance field}

\maketitle
\section{Introduction}

Signed distance fields (SDFs) provide a continuous representation of scene geometry by assigning each spatial location its signed Euclidean distance to the nearest surface \citep{curless1996volumetric}. This representation encodes surfaces as zero-level sets while retaining distance information in the surrounding space. Continuous distance queries, surface reconstruction, and fusion of noisy range measurements make SDFs well-suited to simultaneous localization and mapping (SLAM) and related robotic perception tasks \citep{sh-ch5-map}. In practice, truncated signed distance fields (TSDFs) limit the range over which metric distance information is represented, reducing the storage and computational demands of volumetric mapping \citep{newcombe2011kinectfusion}.

The quality of an SDF map, however, depends on the poses used to integrate observations. In conventional tracking and fusion pipelines \citep{newcombe2011kinectfusion,bylow2013real,kahler2016real}, incoming observations are first aligned to the current map, and the estimated poses are then used to fuse those observations into the field. Although fusion aggregates noisy measurements into a useful geometric reference, integrating observations with imperfect poses can embed pose errors in the reconstructed geometry. Subsequent pose estimation against the current field does not jointly revisit the historical poses and distance values responsible for the distortion. This motivates treating the distance field as an estimation variable coupled to the trajectory, rather than solely as the result of fusion.

\begin{figure}[th!]
\centering
\includegraphics[width=0.485\textwidth]{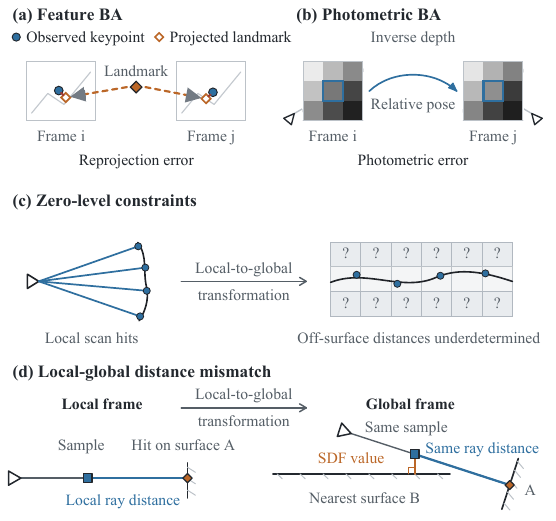}
\caption{Observation--map relationships in conventional BA and the challenges of formulating geometric SDF-BA.
(a) In feature-based BA, reprojection residuals compare observed keypoints with the image projections of their corresponding 3D landmarks.
(b) In photometric BA, photometric residuals compare image intensities in one view with those sampled at locations predicted in another view by the current pose and geometry estimates.
In both cases, a measurement model connects observations to predictions determined by the poses and map.
(c) Surface-hit residuals constrain the SDF zero-level set but do not uniquely determine off-surface metric distances.
(d) For a non-hit sample along the ray, a rigid local-to-global transformation preserves the distance from the sample to the measured hit on surface~A. However, another surface (surface~B in the figure) may be closer to the transformed sample. The signed distance inferred from the ray need not equal the global SDF value at that sample, even when the pose is correct.}
\label{fig:sdf_ba_motivation}
\end{figure}

Joint optimization of poses and maps is a well-established principle in feature-based \citep{konolige2008frameslam,sibley2009adaptive,agarwal2011building,zhao2015parallaxba} and direct SLAM \citep{newcombe2011dtam,engel2017direct}. For example, in visual bundle adjustment (BA), feature-based methods relate image observations to explicit geometric landmarks through reprojection constraints, whereas direct methods jointly optimize poses and scene geometry through photometric consistency, as illustrated in
Figure~\ref{fig:sdf_ba_motivation}(a)--(b).

Conventional feature-based and direct BA benefit from well-established observation models that provide suitable constraints for their corresponding map representations. However, these models cannot be directly applied to other representations. Extending the principle of joint optimization therefore requires representation-compatible observation--map constraints that link measurements to both pose and map variables. A number of attempts have been made in both visual and LiDAR SLAM to address this challenge. For example, surfel-based methods directly constrain surface elements through geometric \citep{cwian2025mad} or photometric residuals \citep{schops2019bad}, while mesh-based approaches couple pose estimation with mesh geometry or deformation
\citep{rosinol2021kimera}. Occupancy-based formulations provide a volumetric alternative by converting range observations into occupied and free space evidence that can be related to the corresponding occupancy map values \citep{wang2025occupancy}. This establishes a direct observation--map relationship through which poses and occupancy variables can be jointly constrained.

For an SDF, however, range observations most directly constrain the zero-level surface rather than the surrounding signed distance values. A measured range endpoint should lie on the zero-level set after transformation into the map frame, as illustrated in Figure~\ref{fig:sdf_ba_motivation}(c). When poses and TSDF values are optimized together, endpoint residuals enforce surface agreement but do not directly specify the signed distance values away from the observed surface. Field deformation can partially compensate for pose inconsistency while maintaining this agreement. Geometric constraints on off-surface values can therefore strengthen the coupling between the trajectory and the surrounding distance field.

Deriving these off-surface constraints from range observations
is nontrivial. Signed distances inferred for non-hit samples
are defined with respect to the locally observed surface and
therefore cannot, in general, be treated directly as measurements of the global SDF. The locally associated surface need not be the globally nearest surface at the transformed sample location. Consequently, the two signed distances may differ even under the correct pose transformation, as illustrated in Figure~\ref{fig:sdf_ba_motivation}(d).

We present SDF-BA, a geometric BA formulation that jointly
optimizes robot poses and explicit TSDF values from LiDAR range observations. Rather than treating locally inferred signed distances as measurements of the global SDF, SDF-BA exploits the geometric relationships among signed distances, field gradients, and surface locations to construct pose-dependent
observation--map constraints. These constraints relate both
measured surface points and off-surface samples along the
observation rays to the global field through the corresponding
poses. They are incorporated into a unified nonlinear least
squares (NLLS) formulation, allowing range observations to jointly constrain the robot trajectory, the zero-level surface, and the surrounding signed distance values within the relevant TSDF region. For large-scale environments, we further integrate SDF-BA into
a submap-based SLAM framework, termed SDF-SLAM. The framework
jointly optimizes robot poses and TSDF values within local
submaps, while global consistency is established by directly aligning overlapping submaps in the SDF domain.

The main contributions of this paper are summarized as follows:
\begin{enumerate}

\item We propose SDF-BA for joint optimization of robot poses and explicit TSDF values from LiDAR range observations. By exploiting the geometric relationships among signed distances, field gradients, surface locations, and ray geometry, SDF-BA establishes pose-dependent constraints on both the zero-level surface and off-surface TSDF values, enabling joint
optimization beyond surface alignment alone.

\item We propose SDF-SLAM, a submap-based framework for large-scale LiDAR SLAM that combines joint optimization of robot poses and TSDF values within local submaps with global optimization of submap poses using direct SDF constraints. Together with a compact TSDF optimization state and sparse map storage, the framework maintains competitive computational efficiency and memory usage compared with other volumetric SLAM approaches.

\item We extensively evaluate SDF-BA and SDF-SLAM on public real-world LiDAR datasets with respect to trajectory accuracy, reconstruction quality, robustness, and computational efficiency. The results show improved trajectory accuracy and reconstruction quality over representative state-of-the-art methods, while ablation studies further demonstrate the benefits of optimizing robot poses and TSDF values together and incorporating geometric constraints beyond the zero-level surface.

\end{enumerate}

\section{Related Work}

\subsection{SDF Mapping}

SDFs represent the environment as a continuous Euclidean distance field, where each spatial location stores the signed distance to its nearest surface \citep{curless1996volumetric}. In practice, TSDFs \citep{newcombe2011kinectfusion}, which store signed distance information only within a truncated region around observed surfaces, are commonly adopted to improve computational efficiency and reduce memory consumption.

When all observations and their corresponding poses are known, SDFs can be constructed by computing the Euclidean distance from each voxel to its nearest surface and determining the sign according to the inside or outside relation. Classical numerical techniques for distance field construction include level-set and fast-marching methods \citep{sethian1996fast,osher2004level}. For online robotic mapping, incremental TSDF methods integrate successive range observations into a global volumetric representation. Following the KinectFusion paradigm \citep{newcombe2011kinectfusion}, signed distance measurements along observation rays are typically combined through weighted fusion, with subsequent methods improving scalability and computational efficiency for robotic sensing. Voxblox \citep{oleynikova_voxblox_2017} introduced an efficient sparse TSDF/ESDF mapping framework for online planning, while nvblox \citep{millane2024nvblox} further exploits GPU acceleration for large-scale real-time volumetric reconstruction. More recently, DB-TSDF \citep{maese2026dbtsdf} introduces a directional bitmask-based TSDF formulation for efficient CPU-based LiDAR volumetric mapping.

Beyond conventional projective TSDF fusion, alternative distance field formulations have been developed to more faithfully approximate Euclidean distance and support efficient spatial queries. FIESTA \citep{han2019fiesta} proposes an incremental distance field framework for online motion planning in large-scale environments. Gaussian-process formulations provide continuous probabilistic distance fields \citep{wu2021faithful}, and have been further extended to incremental metric--semantic mapping \citep{zobeidi2022dense}. VDB-GPDF \citep{wu2024vdb} combines Gaussian-process distance field estimation with a Volumetric Dynamic B+ (VDB) tree structure to enable scalable online mapping while preserving continuous distance and gradient queries.

Neural implicit representations have also been increasingly explored for SDF mapping. iSDF \citep{Ortiz:etal:iSDF2022} incrementally learns a continuous SDF from depth observations. More recent approaches further address the limitations of online neural distance field reconstruction. LGSDF \citep{yue2025lgsdf} combines continual global learning with local SDF updates, while UN3-Mapping \citep{song2025un3} incorporates uncertainty-aware non-projective SDF estimation for 3D reconstruction.

\subsection{SDF-Based Geometric Registration and SLAM}

SDF representations have been widely used for geometric registration and SLAM because their continuous distance values and spatial gradients provide smooth residuals for pose estimation. Rather than relying on explicit feature correspondences, sensor observations can be aligned directly with the geometry implicitly represented by the distance field.

Early SDF-based tracking methods primarily exploited the zero-level surface of the SDF. KinectFusion \citep{newcombe2011kinectfusion}, for example, raycasts the current TSDF model to obtain a predicted surface and estimates the camera pose through coarse-to-fine projective ICP with point-to-plane residuals. InfiniTAM v1 and v2 \citep{kahler2015very,kahler2016real} follow a similar tracking and fusion paradigm while improving the efficiency and scalability of volumetric reconstruction.

Subsequent approaches exploit the distance field more directly for pose estimation. \cite{bylow2013real} minimize signed distance values evaluated at transformed observations and use the corresponding SDF gradients to optimize the camera pose, avoiding explicit surface extraction and conventional correspondence search. EM-Fusion \citep{strecke2019fusion} incorporates robust estimation and probabilistic data association into SDF-based tracking for dynamic scenes. Gradient-SDF \citep{sommer2022gradient} augments the volumetric representation with voxel-wise gradient information to support direct geometric tracking. Related formulations have also been investigated for partial-to-full point cloud registration \citep{li2025partial}, where signed distance values and precomputed gradients provide correspondence-free geometric constraints under partial overlap, noise, and outliers. More recently, closely related distance field formulations have been extended to LiDAR state estimation. D-LIO \citep{cotoelena2026dlio} performs direct LiDAR--inertial odometry against an incrementally maintained truncated distance field, whereas 2Fast-2Lamaa \citep{legentil20242fast2lamaa} leverages continuous distance fields for large-scale LiDAR--inertial SLAM.

Recent neural implicit approaches have further incorporated learned SDF representations into registration and SLAM \citep{pan2024pin,tian2025miso,pan2025rss}. These methods exploit continuously parameterized scene geometry for pose estimation and mapping, and some introduce tighter coupling between trajectory estimation and optimization of the underlying implicit representation. Unlike explicit volumetric SDF formulations, however, the optimized map variables are typically neural features or implicit representation parameters rather than explicit metric signed distance values stored in a volumetric grid.

Beyond scan-to-map alignment, SDFs have also been exploited for registration between volumetric maps. SDF-2-SDF \citep{slavcheva2018sdf} estimates the relative transformation between two SDF representations by minimizing their discrepancy directly in the distance field domain. Voxgraph~\citep{reijgwart2019voxgraph} incorporates registration residuals obtained by evaluating the surface points of one submap in the distance field of another directly into the optimization of
overlapping submap poses.

For geometric registration based on explicit SDF or TSDF representations, the distance field has predominantly served as a geometric reference for estimating poses or relative transformations. Whether registration is performed against the zero-level surface, directly through signed distance residuals, or between overlapping SDF submaps, the geometric information encoded in the current map primarily constrains the transformation variables. The explicit signed distance values themselves, by contrast, are generally not optimized together with the poses from the same multi-view range measurements within the geometric registration objective. The coupling is therefore predominantly from the map to the pose, rather than from the observations jointly to both the pose and the distance field.

This leaves an important gap between SDF-based geometric registration and geometric BA over an explicit metric SDF representation. In particular, how multi-view range observations can jointly constrain robot poses and explicit signed distance values, including both the zero-level surface and the off-surface region, remains largely unexplored.

\subsection{Joint Optimization of Poses and Maps}

Decoupling pose estimation and mapping can reduce the computational complexity of SLAM by allowing the two components to be updated sequentially. In such a formulation, the current pose is typically estimated using the available observations and map, after which the resulting pose estimate is held fixed while the corresponding observations are integrated to update the map. The updated map may subsequently serve as the geometric reference for estimating later poses, but the pose and map variables are not simultaneously adjusted within a common optimization. Consequently, errors in the pose estimate used for map integration can be absorbed into the map representation, while inconsistencies introduced into the map cannot, within the same estimation step, feed back to jointly correct the poses and map variables that produced them \citep{triggs1999bundle}. Although subsequent pose refinement or map reintegration may partially mitigate these errors, the two estimates remain optimized separately rather than being jointly reconciled from the same set of measurement constraints. This separation can limit the achievable consistency and accuracy of both the trajectory and the map.

Joint pose--map optimization addresses this limitation by treating poses and map parameters as coupled variables within a unified objective, such that the same measurement residuals simultaneously constrain and optimize both estimates. This allows inconsistencies to be distributed consistently between the trajectory and the map rather than being absorbed by one while the other is held fixed. In visual SLAM, this principle is most commonly realized through BA, which has been extensively studied in both feature-based and direct formulations.

In feature-based BA, joint optimization is typically formulated over explicit geometric primitives and can be broadly categorized into point-based \citep{triggs1999bundle,konolige2008frameslam,agarwal2011building,zhao2015parallaxba}, line-based \citep{bartoli2005structure,pumarola2017pl}, and plane-based BA \citep{kaess2015simultaneous,liu2023efficient}, depending on the type of features used. In these approaches, sensor observations are associated with corresponding geometric features in the map, allowing measurement residuals to be naturally defined between each observation and its associated map entity. Direct BA similarly performs joint optimization by enforcing photometric consistency between observed image intensities and their predictions at corresponding pixel locations under the current pose and scene estimate \citep{newcombe2011dtam,engel2017direct}. In both cases, the observation--map relationship is explicitly defined by the underlying measurement model, enabling the construction of well-defined BA residuals.

\begin{figure*}[ht!]
\centering
\includegraphics[width=0.95\textwidth]{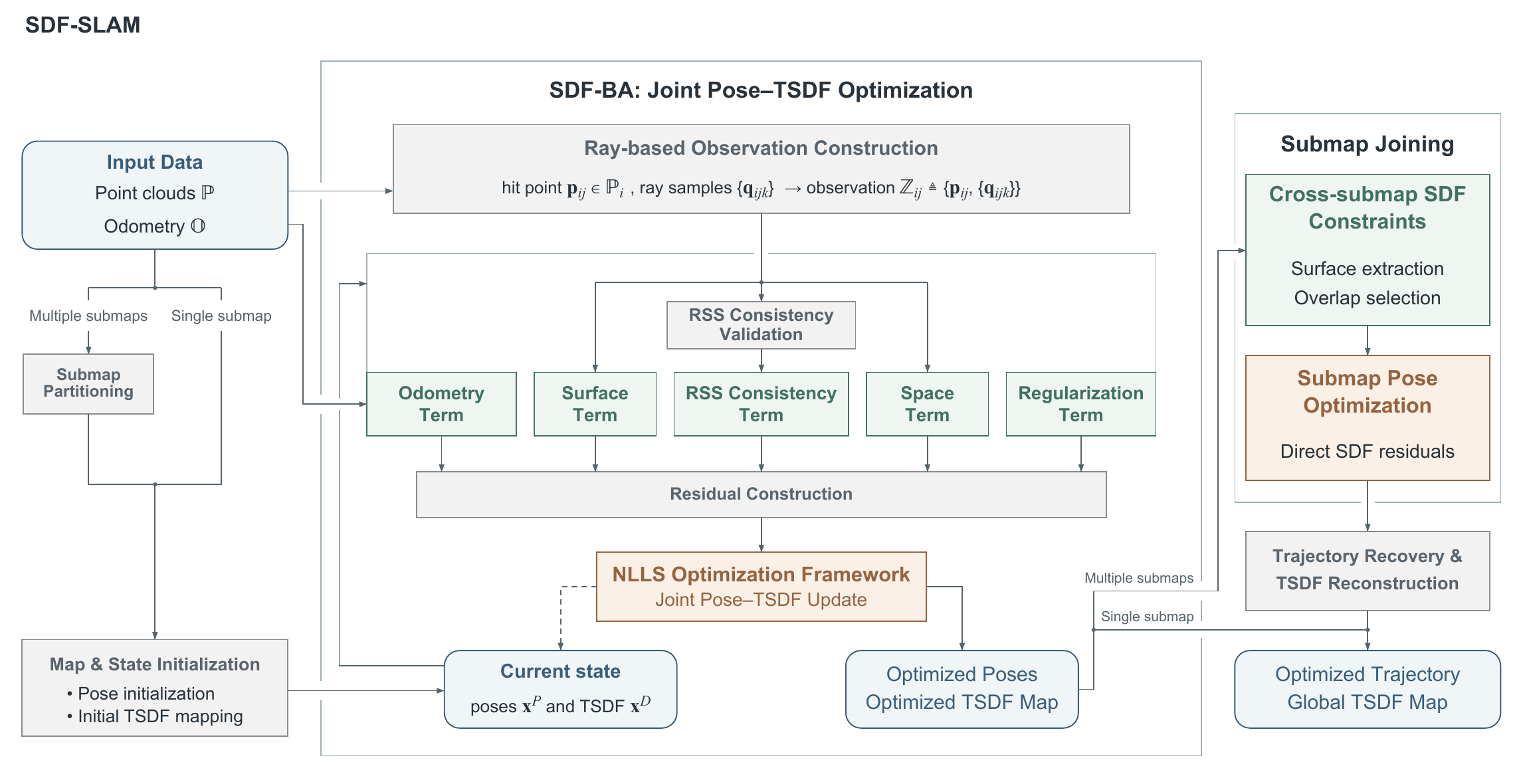}
\caption{\label{fig_overview} Overview of the proposed method. The framework is centered on the proposed SDF-BA, which constructs ray-based observations and jointly optimizes robot poses and the TSDF map within an NLLS framework. For larger environments, the optimized local submaps are further integrated through direct SDF-based submap graph optimization to obtain a globally consistent trajectory and global TSDF map.}
\end{figure*}

Feature-based and photometric BA benefit from well-established observation models that directly couple measurements, poses, and map variables. For dense representations, however, no similarly general observation--pose--map model exists, and the required constraints depend on how the chosen representation parameterizes the map and relates its variables to sensor observations through the poses.

Mesh-based approaches, such as Kimera-PGMO \citep{rosinol2021kimera}, jointly optimize robot poses and mesh deformation by parameterizing the scene geometry through mesh vertices and introducing geometric constraints between poses and vertex positions. Joint optimization over mesh representations has also been investigated from a photometric perspective, where camera poses and mesh vertex intensities are optimized simultaneously \citep{zhang2024direct}. Although formulated on meshes, these approaches retain observation--map relationships closely related to those used in conventional geometric or photometric BA.

A representative surfel-based approach is BAD-SLAM \citep{schops2019bad}, which optimizes camera poses and surfel parameters using geometric and photometric residuals, with poses and surfels updated in an alternating manner for computational efficiency. More recently, MAD-BA \citep{cwian2025mad} introduces a surfel-based BA formulation for LiDAR by jointly optimizing sensor poses and surfel parameters using uncertainty-weighted point-to-plane constraints, while updating data associations during optimization.

Grid-based representations have also been incorporated into joint pose--map optimization, including occupancy grid maps \citep{Zhao-RSS-22,wang2024grid,wang2025occupancy} and depth grid maps \citep{zhou2025correspondence}. Occupancy-SLAM \citep{Zhao-RSS-22,wang2025occupancy} jointly optimizes robot poses and occupancy values by relating sampled observations to the occupancy map, while its submap-joining extension \citep{wang2024grid} jointly optimizes submap poses and a global occupancy map for large-scale mapping. Depth-based formulations \citep{zhou2025correspondence} similarly exploit the direct geometric relationship between spatial observations and depth values to jointly optimize camera poses and depth maps.

Joint optimization of camera poses and SDF values has also
been explored through photometric formulations. Intrinsic3D \citep{maier2017intrinsic3d} jointly optimizes camera parameters, voxel-wise signed distances, and appearance-related scene parameters using photometric and shading cues, while Gradient-SDF \citep{sommer2022gradient} introduces a semi-implicit representation for photometric BA over camera poses and SDF geometry. Although these methods optimize SDF variables, their observation models remain primarily surface-driven: off-surface values are constrained mainly through their influence on the represented surface geometry and through regularization, rather than directly by geometric range observations. Since these values also affect surface locations, normals, and pose updates, errors in the surrounding field can still influence both geometry and camera estimation, leaving the consistency between poses and off-surface signed distances only indirectly constrained.

This distinction becomes particularly relevant when the optimization relies purely on geometric range observations. Unlike photometric consistency, which constrains shared surface geometry through multi-view image observations, an endpoint-to-zero-level constraint requires each transformed range measurement to lie on the estimated surface. Such constraints can provide effective pose information, but when both poses and explicit SDF values are optimized, they leave additional freedom in how pose and map updates explain the same surface observations. This can weaken the coupling between the trajectory and the surrounding distance field, particularly when changes in the field can compensate for part of the pose error. A geometric joint formulation therefore
requires additional relationships that constrain how the trajectory
and distance field are adjusted together, rather than relying on
surface agreement alone.

The ray structure of range observations provides additional geometric information beyond the measured endpoints. Directly exploiting this
information to constrain both surface and off-surface SDF values,
however, requires establishing geometric consistency throughout the
relevant SDF band. This is nontrivial because signed distances
inferred from a locally observed surface do not, in general, coincide
with the corresponding values of the global SDF: the locally
associated surface need not be the globally nearest surface at the
transformed sample location. Establishing geometric observation--map
constraints that account for this local-to-global distance mismatch
therefore remains a key challenge for jointly optimizing poses and
explicit SDF values from range observations.

\section{Preliminaries}

Figure~\ref{fig_overview} provides an overview of the proposed framework. At its core is the proposed SDF-BA, which jointly optimizes robot poses and the TSDF map within a unified formulation. For large-scale environments, this local formulation is further extended to a submap-based SLAM framework, where locally optimized submaps are integrated through direct SDF-based submap graph optimization.

To support the formulation presented in the following sections, we first introduce the notation, the geometric properties of the Euclidean SDF, the proposed observation representation, and the TSDF map representation used in this work. Together, these preliminaries establish the mathematical and structural basis of the proposed method. Throughout the remainder of this paper, the terms SDF and TSDF all refer to Euclidean distance-based definitions.

\subsection{Notation and SDF Definition}

Throughout this paper, unless otherwise stated, we adopt the following typographical conventions. Sets are denoted by blackboard bold symbols, such as $\mathbb{R}$; bold uppercase letters represent matrices; bold lowercase letters denote vectors; and regular lowercase letters indicate scalars.

An SDF is a scalar field that assigns each point in space the signed distance to its closest surface. In 3D, it is defined as
$D(\cdot): \mathbb{R}^3 \rightarrow \mathbb{R}$, with negative and
positive values indicating points inside and outside the surface,
respectively.

\subsection{Geometric Properties of SDFs}

We summarize several fundamental geometric properties of SDFs underlying our formulation, assuming differentiability and a unique closest point on the zero-level set.

\textbf{Property 1: Eikonal (Gradient) Property.}
For a signed Euclidean distance field $D(\cdot)$, the gradient satisfies
\begin{equation}
    \|\nabla D(\mathbf{p})\| = 1,  \label{eq:eikonal_property}
\end{equation}
at differentiable points $\mathbf{p}$.

This property, known as the Eikonal equation \citep{sethian1996fast}, follows directly from the definition of the signed Euclidean distance. The gradient points in the outward unit direction along which the signed distance increases. Under the sign convention used in this paper, $-\nabla D(\mathbf{p})$ therefore points toward the closest surface point when $\mathbf{p}$ lies outside the surface.

\textbf{Property 2: Surface Normal Property.}
For a smooth point $\mathbf{s}$ on the zero-level set of the SDF, i.e., $D(\mathbf{s})=0$, the outward-pointing surface normal is given by
\begin{equation}
    \mathbf{n}_s = \nabla D(\mathbf{s}).
\end{equation}

Since the zero-level set is an isosurface of the signed distance function, its gradient is orthogonal to the surface and coincides with the outward surface normal. Throughout the remainder of this paper, the term surface normal refers to this outward-pointing normal.

Properties 1 and 2 imply several geometric consequences that are central to the proposed formulation. These consequences are stated in the following lemmas, with proofs provided in Appendix \ref{app:sdf_properties}.

\textbf{Lemma 1: Closest-Surface Projection.}
Let $\mathbf{p}$ be a point at which the SDF is differentiable and has a unique closest surface point $\mathbf{s}$ on the zero-level set. Then $\mathbf{s}$ can be computed from the SDF value and gradient at $\mathbf{p}$ as
\begin{equation}
    \mathbf{s} = \mathbf{p} - D(\mathbf{p})\, \nabla D(\mathbf{p}).
\end{equation}

\textbf{Lemma 2: Gradient Sharing.}
Let $\mathbf{s}$ be a smooth surface point with outward unit normal $\mathbf{n}_s$, and let $\mathbf{p}$ lie in a local normal neighborhood where $\mathbf{s}$ is the unique closest surface point to $\mathbf{p}$. Then the SDF gradients at $\mathbf{p}$ and $\mathbf{s}$ are identical:
\begin{equation}
    \nabla D(\mathbf{p}) = \nabla D(\mathbf{s}) = \mathbf{n}_s.
\end{equation}

\subsection{Ray-Based Observation Representation}

Let a sequence of $n+1$ point clouds acquired by a LiDAR be denoted as $\mathbb{P} = \{\mathbb{P}_0,\ldots,\mathbb{P}_n\}$, where each point cloud $\mathbb{P}_i = \{\mathbf{p}_{ij}\}$ is expressed in its corresponding local coordinate frame. Each input point cloud is expressed in its corresponding sensor frame, whose origin coincides with the sensing-ray origin. Any fixed extrinsic calibration between the sensor and the robot body frame is absorbed into the pose variable.

The pose of the $i$-th frame is represented by
\begin{equation}
    \mathbf{T}_i =
    \begin{bmatrix}
        \mathbf{R}_i & \mathbf{t}_i \\
        \mathbf{0}^\top & 1
    \end{bmatrix}
    \in \mathrm{SE}(3),
\end{equation}
where $\mathbf{R}_i \in \mathrm{SO}(3)$ and
$\mathbf{t}_i \in \mathbb{R}^3$ denote the rotation and translation
from the $i$-th sensor frame to the global map frame, respectively.

We construct ray-based observations by sampling both before and beyond each measured endpoint, pairing every sample with its originating endpoint. This pairing preserves the geometric relationship between the off-surface samples and the corresponding surface observation.

\begin{figure}[tp]
\centering
\includegraphics[width=0.485\textwidth]{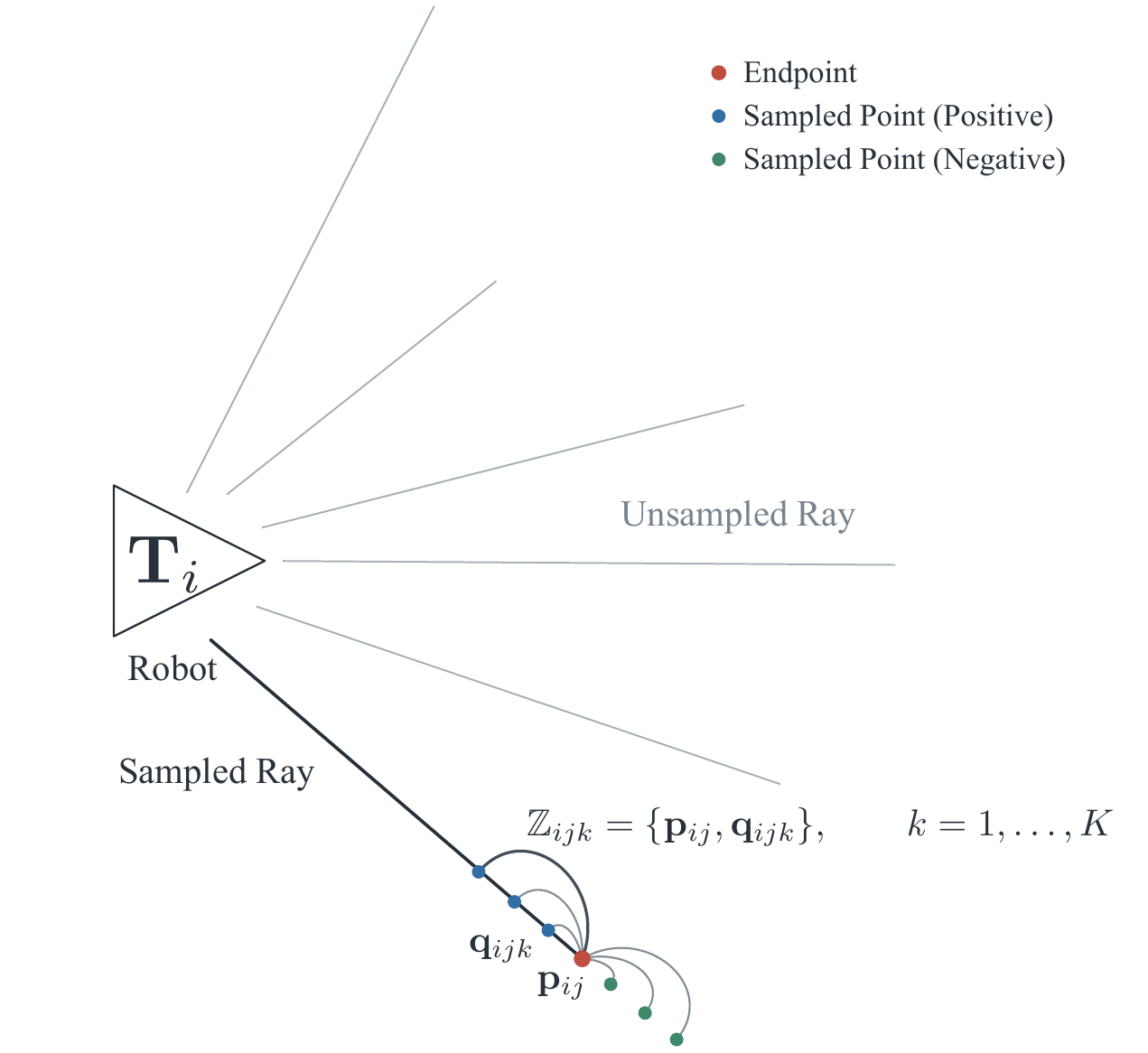}
\caption{{Equidistant sampling strategy for generating observations from a LiDAR scan.}}
\label{fig_sampling}
\end{figure}

Specifically, each point $\mathbf{p}_{ij}\in \mathbb{P}_i$ is regarded as the endpoint (hit point) of a ray originating from the sensor, i.e., from the local frame origin. Along each ray, we perform equidistant sampling with spacing $\Delta s$ over the interval $[-\tau_d,\tau_d]$ centered at $\mathbf{p}_{ij}$, where $\tau_d$ denotes the sampling extent on either side of the endpoint, as illustrated in Figure~\ref{fig_sampling}. Sampling is carried out both before and beyond the measured endpoint along the ray, resulting in a set of $K = 2\tau_d/\Delta s$ sampled points $\{\mathbf{q}_{ijk}\}$. We define the observation associated with the $j$-th ray in the $i$-th frame as a ray-wise geometric observation set
\begin{equation}
\mathbb{Z}_{ij} \triangleq \left\{\mathbf{p}_{ij}, \{\mathbf{q}_{ijk}\}_{k=1}^{K}\right\},
\end{equation}
which serves as the fundamental observation unit in the proposed formulation and encodes the geometric configuration of samples along a single ray. The frame-wise and complete observation sets are defined as
\begin{equation}
    \mathbb{Z}_i
    \triangleq
    \bigcup_{j=1}^{|\mathbb{P}_i|}\mathbb{Z}_{ij},
    \qquad
    \mathbb{Z}
    \triangleq
    \bigcup_{i=0}^{n}\mathbb{Z}_i,
\end{equation}
where $|\mathbb{P}_i|$ denotes the number of points in the
$i$-th point cloud $\mathbb{P}_i$.

For notational convenience, we further define a sample-wise paired observation element derived from $\mathbb{Z}_{ij}$ as
\begin{equation}
\mathbb{Z}_{ijk} \triangleq \left\{\mathbf{p}_{ij}, \mathbf{q}_{ijk}\right\}, \quad k = 1,\ldots,K,
\end{equation}
which pairs the hit point with an individual sampled point along the ray, as shown in Figure~\ref{fig_sampling}.

\subsection{TSDF Map Representation}

The continuous SDF is represented by a voxelized TSDF with voxel
spacing $h$ and truncation distance $\tau$. The optimization
variables are metric signed distance values stored at the voxel
centers, denoted by $D_c \triangleq D(\mathbf{m}_c) \in [-\tau,\tau]$, where $\mathbf{m}_c$ is the center of voxel $c$. Given a sequence of point clouds $\mathbb{P}$ and their associated initial poses $\{\mathbf{T}_i\}$, the discrete TSDF values ${D_c}$ are initialized through volumetric fusion following a paradigm similar to KinectFusion
\citep{newcombe2011kinectfusion}.

For an arbitrary query point $\mathbf{r}$ inside a voxel cell, the continuous field value is obtained by trilinear interpolation,
\begin{equation}
    D(\mathbf{r})
    =
    \sum_{c\in\mathcal{V}(\mathbf{r})}
    \alpha_c(\mathbf{r})D_c,
    \qquad
    \sum_{c\in\mathcal{V}(\mathbf{r})}\alpha_c(\mathbf{r})=1,
    \label{eq:tsdf_trilinear_interpolation}
\end{equation}
where $\mathcal{V}(\mathbf{r})$ indexes the eight neighboring voxel
centers that define the interpolation cell, and $\alpha_c(\mathbf{r})$ is the corresponding interpolation weight. The field gradient is evaluated consistently from the discrete TSDF. Because a TSDF only approximates an exact Euclidean SDF, we explicitly use the normalized gradient direction
\begin{equation}
    \widehat{\mathbf{n}}_D(\mathbf{r})
    =
    \frac{\nabla D(\mathbf{r})}
    {\max(\|\nabla D(\mathbf{r})\|,\epsilon_g)},
    \label{eq:normalized_tsdf_gradient}
\end{equation}
where $\epsilon_g>0$ prevents division by a numerically vanishing gradient. The corresponding discrete closest-surface projection is
\begin{equation}
    \pi_D(\mathbf{r})
    =
    \mathbf{r}-D(\mathbf{r})\widehat{\mathbf{n}}_D(\mathbf{r}).
    \label{eq:discrete_surface_projection}
\end{equation}
For an exact differentiable SDF in a local normal neighborhood, (\ref{eq:discrete_surface_projection}) reduces to Lemma~1.

\section{SDF-BA}
In this section, we present our core component, termed SDF-BA, which jointly optimizes robot poses and a TSDF map within an NLLS framework. We introduce the state representation, the formulation built on the intrinsic properties of the SDF and ray-structured observations, and the corresponding iterative optimization algorithm.

\subsection{State Vector}

The proposed method jointly optimizes the robot poses and a selected set of TSDF values. The state vector is defined as
\begin{equation}
    \mathbf{x}
    =
    \left[
        (\mathbf{x}^P)^\top,
        (\mathbf{x}^D)^\top
    \right]^\top,
\end{equation}
where the pose-related state vector stacks the parameterizations of
$\{\mathbf{T}_i\}_{i=0}^{n}$ as
\begin{equation}
    \mathbf{x}^P
    =
    \left[
        (\mathbf{x}^P_0)^\top,
        \ldots,
        (\mathbf{x}^P_n)^\top
    \right]^\top.
\end{equation}
Each pose state is parameterized as
\begin{equation}
    \mathbf{x}^P_i
    =
    \left[
        \mathbf{t}_i^\top,
        \boldsymbol{\xi}_i^\top
    \right]^\top
    \in \mathbb{R}^6,
\end{equation}
where $\boldsymbol{\xi}_i\in\mathbb{R}^3$ is the rotation vector satisfying
$\mathbf{R}_i=\exp(\boldsymbol{\xi}_i^\wedge)$, with
$\boldsymbol{\xi}_i^\wedge\in\mathfrak{so}(3)$ denoting the corresponding
skew-symmetric matrix. The first pose is fixed throughout optimization to remove the global rigid-body gauge freedom.

The map-related state vector is defined as
\begin{equation}
    \mathbf{x}^D
    =
    \left[
        D(\mathbf{m}_c)
    \right]_{c\in\Omega_D},
    \label{eq_map_state}
\end{equation}
where $\Omega_D$ denotes the fixed set of TSDF voxels included in the optimization. A voxel is included in $\Omega_D$ only if it lies within the valid truncation band and satisfies the required observation and connectivity conditions. Voxels outside $\Omega_D$ that are required for interpolation or gradient computation are retained as fixed query support, but they do not introduce additional optimization degrees of freedom. This preserves the required TSDF query support while keeping the optimized map state compact.

\subsection{NLLS Formulation} \label{sec_NLLS}

The objective function of the {NLLS} problem is defined as
\begin{equation}
\begin{aligned}
    f(\mathbf{x}) ={}&
    w_{0}f^{0}(\mathbf{x})
    + w_{S}f^{S}(\mathbf{x})
    + w_{C}f^{C}(\mathbf{x}) \\
    &{}+ w_{R}f^{R}(\mathbf{x})
    + w_{O}f^{O}(\mathbf{x}).
\end{aligned}
\label{eq_objective}
\end{equation}
The objective function consists of the surface term $f^{0}(\mathbf{x})$, the space term $f^{S}(\mathbf{x})$, the ray-surface-space consistency term $f^{C}(\mathbf{x})$, the regularization term $f^{R}(\mathbf{x})$, and the odometry term $f^O(\mathbf{x})$. $w_0$ , $w_S$, $w_{C}$, $w_R$, and $w_O$ are their corresponding weights. We next explain the five terms one by one.

\textbf{\textit{1) Surface Term:}} The most straightforward constraint on the TSDF is that the transformed measurement endpoint
\begin{equation}
    \mathbf{p}_{ij}^\prime = \mathbf{R}_i\mathbf{p}_{ij}+\mathbf{t}_{i} \label{eq_p_proj}
\end{equation}
should lie on the surface. In other words, its signed distance value $D(\mathbf{p}_{ij}^\prime)$ should be close to zero.

Accordingly, the surface term in (\ref{eq_objective}) is defined as
\begin{equation}
    f^0(\mathbf{x})
    =
    \sum_{i=0}^{n}
    \sum_{j}
    \rho\!\left(
        \left|D(\mathbf{p}_{ij}^{\prime})\right|^2
    \right),
    \label{eq_obj_surface}
\end{equation}
where $\rho(\cdot)$ denotes a robust kernel used to reduce the influence of outlier measurements. This term directly couples the robot poses with the TSDF values and constrains the reconstructed zero-level surface to remain consistent with the observed hit points. However, as illustrated in Figure~\ref{fig:sdf_ba_motivation}(c), the surface term only requires the transformed endpoints to lie on the zero-level set and therefore does not directly constrain the surrounding off-surface TSDF values, allowing different local field configurations to satisfy the same surface observations.

\textbf{\textit{2) Space Term:}} To constrain the field beyond the observed zero-level surface, we introduce a space term over the active TSDF region.

For a voxel $c\in\Omega_D$ whose center lies in a valid local normal neighborhood, the discrete projection in (\ref{eq:discrete_surface_projection}) gives the associated surface estimate
\begin{equation}
    \mathbf{s}_{c}
    = \pi_D(\mathbf{m}_c)
    = \mathbf{m}_{c} - D_c\widehat{\mathbf{n}}_D(\mathbf{m}_{c}).
    \label{eq_surface_found}
\end{equation}

The space term penalizes the TSDF value at this projected estimate:
\begin{equation}
\begin{aligned}
    f^{S}(\mathbf{x})
    &= \sum_{c\in \Omega_D} \left\|D(\mathbf{s}_{c})\right\|^2 \\
    &= \sum_{c\in \Omega_D}
    \left\|D\!\left(\mathbf{m}_{c}
    -D_c\widehat{\mathbf{n}}_D(\mathbf{m}_{c})\right)\right\|^2.
    \end{aligned} \label{eq_obj_space}
\end{equation}

This term penalizes disagreement between each active distance value
and the zero-level location implied by the current field. It extends the zero-level constraint in Figure~\ref{fig:sdf_ba_motivation}(c) to the surrounding field through the intrinsic geometry of the SDF. However, the space term involves only the TSDF and does not directly couple these off-surface values with the range observations and robot poses.

\textbf{\textit{3) Ray-Surface-Space Consistency Term:}} To directly relate off-surface TSDF values to the range observations through the robot poses, we introduce the ray-surface-space (RSS) consistency term. It exploits the geometric relationship between sampled points along each ray and their corresponding endpoints, together with the surface implied by the current TSDF, to jointly constrain the poses and the surrounding distance field.

For each paired observation $\mathbb{Z}_{ijk}=\{\mathbf{p}_{ij},\mathbf{q}_{ijk}\}$, the sample--endpoint pairing is preserved under rigid-body transformations. Specifically,
\begin{equation}
    \mathbf{q}_{ijk}^\prime = \mathbf{R}_i\mathbf{q}_{ijk}+\mathbf{t}_{i}, \label{eq_q_proj}
\end{equation}
where the transformed point $\mathbf{q}_{ijk}^\prime$ remains paired with the corresponding transformed endpoint $\mathbf{p}_{ij}^\prime$ defined in (\ref{eq_p_proj}). However, the preserved sample--endpoint geometry does not imply that their distance equals the global SDF value at $\mathbf{q}_{ijk}^\prime$. As illustrated in Figure~\ref{fig:sdf_ba_motivation}(d), another surface may be closer to $\mathbf{q}_{ijk}^\prime$ than the surface observed at the paired endpoint.

\begin{figure}[tp]
\centering
\includegraphics[width=0.485\textwidth]{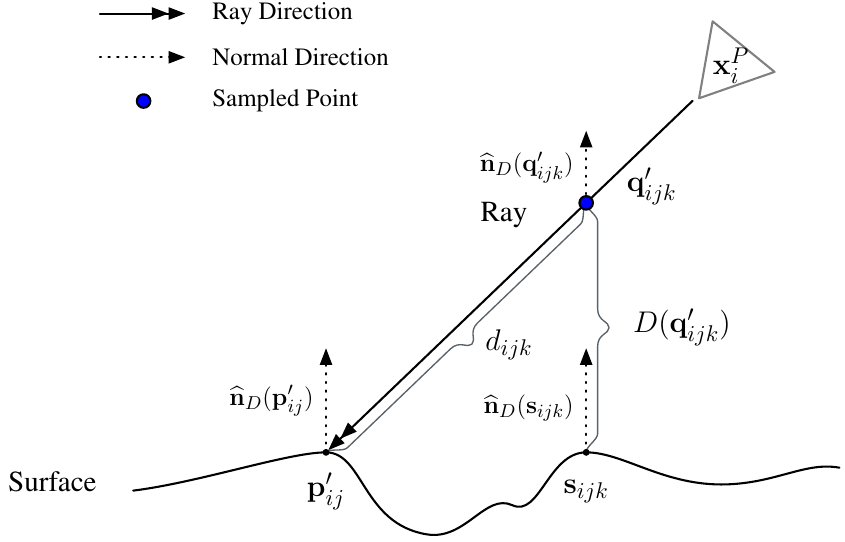}
\caption[Geometric interpretation of the ray-surface-space consistency constraint]
{Geometric interpretation of the ray-surface-space consistency constraint. The sampled point $\mathbf{q}_{ijk}^\prime$ is projected to the surface estimate $\mathbf{s}_{ijk}$ using the TSDF value $D(\mathbf{q}_{ijk}^\prime)$ and the normalized field gradient. The corresponding ray endpoint is
$\mathbf{p}^\prime_{ij}$, and the ray direction in the global frame is
$\mathbf{R}_i\mathbf{n}_{ij}$.}
\label{fig_RSS}
\end{figure}

To address this limitation, the RSS consistency term combines the preserved sample--endpoint geometry with the local geometry inferred from the current TSDF, rather than equating the sample--endpoint distance with the global SDF, as illustrated in Figure~\ref{fig_RSS}.

Specifically, the surface estimate $\mathbf{s}_{ijk}$ associated with
$\mathbf{q}_{ijk}^\prime$ is computed from the current TSDF using
the projection in \eqref{eq:discrete_surface_projection}:
\begin{equation}
    \mathbf{s}_{ijk}
    = \pi_D(\mathbf{q}_{ijk}^\prime)
    = \mathbf{q}^\prime_{ijk}
    -D(\mathbf{q}^\prime_{ijk})
    \widehat{\mathbf{n}}_D(\mathbf{q}^\prime_{ijk}).
    \label{eq_q_surface_found}
\end{equation}
This estimate is obtained from local TSDF values and gradients
rather than by searching for a corresponding discrete surface
element.

Lemma~2 motivates a shared-normal approximation between
$\mathbf{q}_{ijk}^{\prime}$ and its associated surface estimate
$\mathbf{s}_{ijk}$. If the local normal at the ray endpoint
$\mathbf{p}_{ij}^{\prime}$ is also consistent with that at
$\mathbf{s}_{ijk}$, we obtain
\begin{equation}
    \widehat{\mathbf{n}}_D(\mathbf{p}_{ij}^\prime)
    \simeq
    \widehat{\mathbf{n}}_D(\mathbf{s}_{ijk})
    \simeq
    \widehat{\mathbf{n}}_D(\mathbf{q}_{ijk}^\prime).
    \label{eq_shared_normal}
\end{equation}
We denote this shared local surface normal by $\mathbf{n}_s$.

For a pair $(\mathbf{p}_{ij}^\prime,\mathbf{s}_{ijk})$ satisfying
the geometric compatibility conditions described in
Section~\ref{sec:rss_consistency_filtering}, the scalar geometric
residual is defined as
\begin{equation}
    r^{C}_{ijk}
    =
    {\mathbf{n}_s}^\top
    \left(
        \mathbf{s}_{ijk}
        -
        \mathbf{p}_{ij}^\prime
    \right).
    \label{eq_plane}
\end{equation}
The required geometric compatibility is not restricted to planar or coplanar configurations, and RSS can also accommodate locally smooth curved surfaces that satisfy the same compatibility conditions. This distinguishes RSS from conventional plane-based methods that rely on explicit plane extraction and coplanarity constraints.

Substituting (\ref{eq_q_surface_found}) into (\ref{eq_plane}) and applying the shared-normal approximation in (\ref{eq_shared_normal}), the RSS residual used in the optimization
is written as
\begin{equation}
    r^{C}_{ijk}(\mathbf{x})
    =
    -d_{ijk}\mathbf{n}_s^\top\mathbf{R}_i\mathbf{n}_{ij}
    -
    D(\mathbf{q}_{ijk}^{\prime}),
    \label{eq_RSS_formal}
\end{equation}
where
\begin{equation}
    d_{ijk}
    =
    \operatorname{sign}
    \left(
        \lVert \mathbf{p}_{ij}\rVert
        -
        \lVert \mathbf{q}_{ijk}\rVert
    \right)
    \lVert
        \mathbf{p}_{ij}
        -
        \mathbf{q}_{ijk}
    \rVert
    \label{eq_local_dist}
\end{equation}
denotes the signed offset from $\mathbf{q}_{ijk}$ to its paired
endpoint $\mathbf{p}_{ij}$ along the ray direction, and
\begin{equation}
    \mathbf{n}_{ij}
    =
    \frac{\mathbf{p}_{ij}}
    {\lVert \mathbf{p}_{ij}\rVert}
\end{equation}
is the corresponding ray direction in the $i$-th local frame.

The RSS consistency term is then defined as
\begin{equation}
    f^{C}(\mathbf{x})
    =
    \sum_{i=0}^{n}
    \sum_{(j,k)\in\Omega_C^i}
    \rho_C
    \left(
        \left|r^{C}_{ijk}(\mathbf{x})\right|^2
    \right),
    \label{eq:rss_discrete_energy}
\end{equation}
where $\Omega_C^i$ denotes the index set of valid RSS sample pairs at timestamp $i$, and $\rho_C(\cdot)$ denotes a robust kernel.

Unlike the surface term, which evaluates the TSDF at measured
endpoints, the RSS term evaluates it at paired off-surface samples
and relates these values to the corresponding endpoints through
the ray geometry and local field normal in
(\ref{eq_RSS_formal}). It therefore provides observation-derived
constraints on the TSDF values supporting these queries while
jointly constraining the corresponding poses. This complements
the space term, which enforces consistency between the field
values and their projected zero-level points.

\textbf{\textit{4) Regularization Term:}} The observation-derived terms do not constrain all spatial frequencies of the active TSDF equally. We therefore use a discrete Laplacian to couple neighboring map states and suppress unsupported high-frequency variation without anchoring the map to its initial values. Let $\mathcal{N}_c\subset\Omega_D$ be the available 6-connected neighbors of voxel $c$ and let $\phi_c=|\mathcal{N}_c|$. The regularization term is
\begin{equation}
    f^R(\mathbf{x})
    =\sum_{c\in\Omega_D}
    \left\|
    \phi_cD_c-\sum_{l\in\mathcal{N}_c}D_l
    \right\|^2.
    \label{eq:laplacian_regularization}
\end{equation}

\textbf{\textit{5) Odometry Term:}} When available, the odometry measurements $\mathbb{O}=\{\overline{\mathbf{T}}_{i-1,i}\}_{i=1}^{n}$ provide relative transformations between consecutive robot poses. The residual is defined directly on $\mathrm{SE}(3)$ as
\begin{equation}
    \mathbf r_i^O(\mathbf{x})
    =
    \operatorname{Log}\!\left(
    \overline{\mathbf{T}}_{i-1,i}^{-1}
    \mathbf{T}_{i-1}^{-1}\mathbf{T}_{i}
    \right)\in\mathbb{R}^6,
    \label{eq:odom_se3_residual}
\end{equation}
and
\begin{equation}
    f^O(\mathbf{x})
    =
    \sum_{i=1}^{n}
    \left\|\mathbf r_i^O(\mathbf{x})\right\|_{\mathbf\Sigma_{O_i}^{-1}}^2,
    \label{eq:odom_energy}
\end{equation}
where $\operatorname{Log}:\mathrm{SE}(3)\rightarrow\mathbb{R}^6$ is the logarithm map and $\mathbf\Sigma_{O_i}$ is the covariance of the relative-pose measurement.

\subsection{RSS Consistency Validation}\label{sec:rss_consistency_filtering}

\begin{figure}[tp]
\centering
\includegraphics[width=0.485\textwidth]{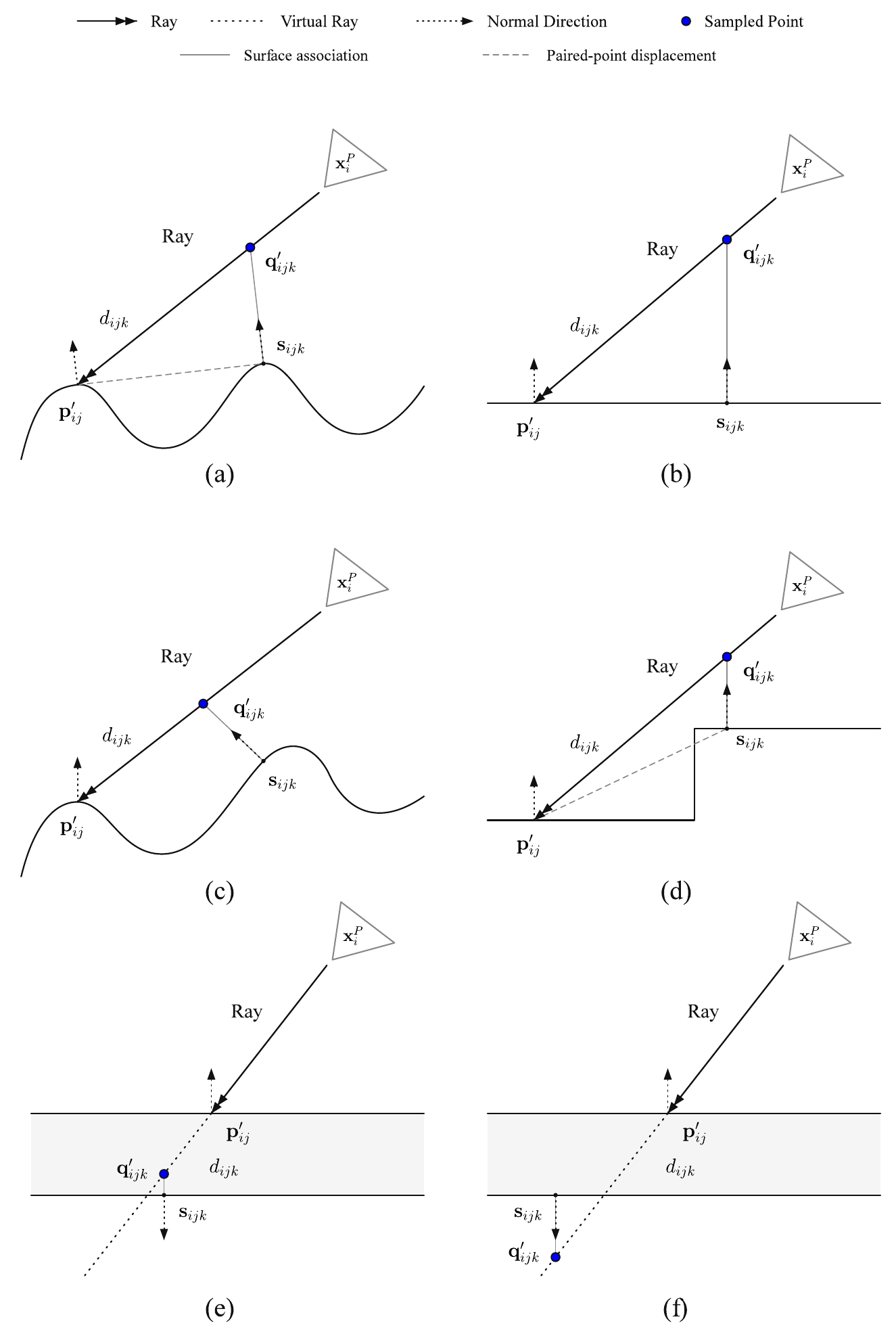}
\caption[Different cases of RSS consistency checks]
{Examples of RSS consistency checks.
Cases (a) and (b) illustrate valid curved-surface and planar
configurations, respectively. Case (c) fails the normal
consistency check, while case (d) fails the line--normal
orthogonality check. Cases (e) and (f) illustrate virtual-ray
configurations in which the sampled point becomes associated with
a different surface region and fails the normal consistency check.
Pairs failing either check are excluded from the RSS consistency term.}
\label{fig_RSS_case}
\end{figure}

For each paired ray sample, the RSS consistency term relates the transformed endpoint
$\mathbf{p}_{ij}^\prime$ to the surface estimate $\mathbf{s}_{ijk}$
obtained by projecting $\mathbf{q}_{ijk}^\prime$ through the current
TSDF. Due to the local-to-global ambiguity illustrated in
Figure~\ref{fig:sdf_ba_motivation}(d), these two locations may correspond to different local surface regions in geometrically complex environments, as further illustrated in Figure~\ref{fig_RSS_case}. We therefore validate each pair using
geometric consistency and proximity checks before including it in the RSS consistency term.

\textbf{1) Normal Consistency}: To satisfy the shared-normal
approximation in (\ref{eq_shared_normal}), the local normals at
$\mathbf{p}_{ij}^\prime$ and $\mathbf{s}_{ijk}$ should be sufficiently
consistent. The pair is retained only if
\begin{equation}
    \widehat{\mathbf{n}}_D(\mathbf{p}_{ij}^\prime)^\top
    \widehat{\mathbf{n}}_D(\mathbf{s}_{ijk})
    \geq
    \cos(\tau_{\mathrm{norm}}).
    \label{eq:rss_normal_gate}
\end{equation}

\textbf{2) Line--Normal Orthogonality}: For a normal-consistent pair,
define the averaged normal as
\begin{equation}
    \bar{\mathbf{n}}_{sp}
    =
    \frac{
        \widehat{\mathbf{n}}_D(\mathbf{p}_{ij}^\prime)
        +
        \widehat{\mathbf{n}}_D(\mathbf{s}_{ijk})
    }{
        \left\|
        \widehat{\mathbf{n}}_D(\mathbf{p}_{ij}^\prime)
        +
        \widehat{\mathbf{n}}_D(\mathbf{s}_{ijk})
        \right\|
    }.
    \label{eq_avg_normal}
\end{equation}
To satisfy the geometric compatibility underlying
(\ref{eq_plane}), the displacement between the endpoint and its
associated surface estimate should be approximately orthogonal to
their averaged local normal:
\begin{equation}
    \left|
    \bar{\mathbf{n}}_{sp}^\top
    (\mathbf{p}_{ij}^\prime-\mathbf{s}_{ijk})
    \right|
    <
    \tau_{\mathrm{orth}}.
    \label{eq:rss_orthogonal_gate}
\end{equation}

As illustrated in Figure~\ref{fig_RSS_case}, cases~(a) and~(b)
satisfy both geometric consistency checks and illustrate curved-surface
and planar RSS configurations, respectively. Case~(c) fails the normal
consistency check, while case~(d) fails the line--normal orthogonality
check. Cases~(e) and~(f) further illustrate situations in which samples
on the virtual continuation of a ray become associated with different
surface regions beyond the first surface intersection and fail the
normal consistency check. Pairs failing either geometric check do not contribute to the RSS consistency term. Therefore, the physical thickness
of an obstacle need not be known when generating the ray samples.

Pairs satisfying both geometric checks proceed to the proximity
validation. The remaining pairs also require reliable support from the current discrete TSDF. Pairs whose sampled points lack a complete valid interpolation stencil or have a numerically vanishing gradient are excluded before the
following checks.

\textbf{3) Surface Proximity}: The projected surface point
$\mathbf{s}_{ijk}$ computed using (\ref{eq_q_surface_found}) should
lie sufficiently close to the zero-level surface. The pair is retained
only if
\begin{equation}
    |D(\mathbf{s}_{ijk})|
    <
    \tau_{\mathrm{surf}}.
    \label{eq:rss_surface_proximity}
\end{equation}

\textbf{4) Spatial Proximity}: We further examine whether the spatial
separation between $\mathbf{s}_{ijk}$ and $\mathbf{p}_{ij}^\prime$
is consistent with the sampled ray geometry. Using the averaged normal
$\bar{\mathbf{n}}_{sp}$ defined in (\ref{eq_avg_normal}), the pair is
retained only if
\begin{equation}
    \left|
    |d_{ijk}|
    \left\|
        \bar{\mathbf{n}}_{sp}
        \times
        (\mathbf{R}_i\mathbf{n}_{ij})
    \right\|
    -
    \|\mathbf{s}_{ijk}-\mathbf{p}_{ij}^\prime\|
    \right|
    <
    \tau_{\mathrm{dist}}.
    \label{eq:rss_spatial_gate}
\end{equation}

The four checks define the index set $\Omega_C^i$ of valid RSS sample pairs, which is recomputed at each nonlinear iteration from the current pose and TSDF estimates. This validation corresponds to Line~\ref{line:rss_validation} in Algorithm~\ref{alg_flowchart}.

\begin{algorithm}[t]
\caption{SDF-BA Algorithm}
\label{alg_flowchart}
\SetKwInput{KwParam}{Params}
\SetKwInput{KwInput}{Input}
\SetKwInput{KwOutput}{Output}
\small
\SetAlgoLined
\DontPrintSemicolon

\SetKwFunction{FuncReturn}{return}
\SetKwFunction{FuncDown}{Sampling}
\SetKwFunction{FuncInitPose}{Pose Initialization}
\SetKwFunction{FuncInitMap}{TSDF Initialization}
\SetKwFunction{FuncSelectMap}{Map-State Selection}
\SetKwFunction{FuncUpdateField}{TSDF Update}
\SetKwFunction{FuncProj}{Transformation}
\SetKwFunction{FuncAssociate}{Surface Projection}
\SetKwFunction{FuncRSSValidate}{RSS Consistency Validation}
\SetKwFunction{FuncLinearize}{Linearization}
\SetKwFunction{FuncDelta}{Step Solve}
\SetKwFunction{FuncConverge}{Convergence Test}

\KwParam{Sampling parameter $\tau_d$, maximum iteration number $k_{\max}$,
convergence threshold $\tau_{\Delta}$, geometric thresholds
$\tau_{\mathrm{norm}}$, $\tau_{\mathrm{orth}}$, $\tau_{\mathrm{dist}}$,
and $\tau_{\mathrm{surf}}$, and weights $w_0$, $w_S$, $w_C$, $w_R$,
and $w_O$}
\KwInput{Point clouds $\mathbb{P}$ and odometry $\mathbb{O}$}
\KwOutput{Optimized poses $\hat{\mathbf{x}}^P$ and map states
$\hat{\mathbf{x}}^D$}

$\mathbb{Z} \leftarrow$ \FuncDown{$\mathbb{P}$, $\tau_d$}\;

$\mathbf{x}^P(0) \leftarrow$ \FuncInitPose{$\mathbb{O}$}\;

$D(\cdot)
\leftarrow$
\FuncInitMap{$\mathbf{x}^P(0)$, $\mathbb{Z}$}\;

$\Omega_D, \mathbf{x}^D(0)
\leftarrow$
\FuncSelectMap{$D(\cdot)$, $\mathbb{Z}$}\;

$\mathbf{x}(0)
\leftarrow
\left[
\mathbf{x}^P(0)^\top,
\mathbf{x}^D(0)^\top
\right]^\top$\;

$k \leftarrow 0$, $converged \leftarrow \mathrm{false}$\;

\While{$k < k_{\max}$ and \textbf{not} $converged$}{

    $D(\cdot)
    \leftarrow$
    \FuncUpdateField{$\mathbf{x}^D(k)$, $\Omega_D$}\;

    \For{$i=0,\ldots,n$}{

        $\{\mathbf{p}'_{ij},\mathbf{q}'_{ijk}\}
        \leftarrow$
        \FuncProj{$\mathbf{x}^P_i(k)$, $\mathbb{Z}_i$}\;

        $\{\mathbf{s}_{ijk}\}
        \leftarrow$
        \FuncAssociate{$D(\cdot)$, $\{\mathbf{q}'_{ijk}\}$}\;

        $\Omega_C^i(k)
        \leftarrow$
        \FuncRSSValidate{
            $D(\cdot)$,
            $\{(\mathbf{p}'_{ij},\mathbf{q}'_{ijk},\mathbf{s}_{ijk})\}$,
            $\tau_{\mathrm{norm}}$,
            $\tau_{\mathrm{orth}}$,
            $\tau_{\mathrm{surf}}$,
            $\tau_{\mathrm{dist}}$
        }\;
        \label{line:rss_validation}
    }

    $F(\mathbf{x}(k)),\mathbf{J}(k)
    \leftarrow$
    \FuncLinearize{
        $\mathbb{Z}$,
        $\mathbb{O}$,
        $\mathbf{x}(k)$,
        $\{\Omega_C^i(k)\}_{i=0}^{n}$
    }\;

    $\boldsymbol{\Delta}(k)
    \leftarrow$
    \FuncDelta{
        $F(\mathbf{x}(k))$,
        $\mathbf{J}(k)$,
        $\mathbf{W}$
    }\;

    $\mathbf{x}(k+1)
    \leftarrow
    \mathbf{x}(k)\boxplus\boldsymbol{\Delta}(k)$\;

    $converged
    \leftarrow$
    \FuncConverge{
        $\boldsymbol{\Delta}(k)$,
        $\tau_{\Delta}$
    }\;

    $k \leftarrow k+1$\;
}

$\hat{\mathbf{x}}^P
\leftarrow
\mathbf{x}^P(k)$,
$\hat{\mathbf{x}}^D
\leftarrow
\mathbf{x}^D(k)$\;

\FuncReturn{$\hat{\mathbf{x}}^P$, $\hat{\mathbf{x}}^D$}\;

\end{algorithm}

\subsection{Optimization Procedure and Solution}

The proposed SDF-BA jointly estimates the robot poses and TSDF voxel
values by minimizing the nonlinear objective defined in
(\ref{eq_objective}), i.e.,
\begin{equation}
    \hat{\mathbf{x}}
    =
    \arg\min_{\mathbf{x}} f(\mathbf{x}).
    \label{eq_sdfba_optimization}
\end{equation}
Since the residuals depend nonlinearly on the pose variables, the
problem is solved iteratively from an initial estimate
$\mathbf{x}(0)$.

Let
$F^0(\mathbf{x})$, $F^S(\mathbf{x})$, $F^C(\mathbf{x})$,
$F^R(\mathbf{x})$, and $F^O(\mathbf{x})$
denote the residual vectors associated with the surface, space, RSS
consistency, regularization, and odometry terms, respectively. They
are stacked as
\begin{equation}
\scalebox{0.99}{$
    \displaystyle
    F(\mathbf{x})
    =
    \left[
        F^0(\mathbf{x})^\top,
        F^S(\mathbf{x})^\top,
        F^C(\mathbf{x})^\top,
        F^R(\mathbf{x})^\top,
        F^O(\mathbf{x})^\top
    \right]^\top
$}.
\end{equation}
The corresponding block-diagonal weight matrix $\mathbf{W}$ collects
the weights of the individual objective terms, with the odometry
residuals additionally weighted by their inverse covariance matrices.
Robust kernels are incorporated through the corresponding residual
weights during each linearization.

At iteration $k$, the valid RSS sample pairs are first determined from
the current pose and TSDF estimates and are held fixed while the
current local problem is linearized and solved. With a local increment
$\boldsymbol{\Delta}$, the residual vector is approximated as
\begin{equation}
    F\!\left(
        \mathbf{x}(k)\boxplus\boldsymbol{\Delta}
    \right)
    \approx
    F(\mathbf{x}(k))
    +
    \mathbf{J}(k)\boldsymbol{\Delta},
    \label{eq_local_linearization}
\end{equation}
where $\boxplus$ applies a local Lie-group increment to the pose
variables and additive increments to the TSDF variables, and
$\mathbf{J}(k)$ is the analytic Jacobian with respect to the local
increment. The derivation of the RSS consistency residual
Jacobian is provided in Appendix \ref{app:jacobians}.

The local increment is obtained by solving the damped normal equations
\begin{equation}
    \left(
        \mathbf{J}^\top\mathbf{W}\mathbf{J}
        +
        \lambda\mathbf{I}
    \right)
    \boldsymbol{\Delta}
    =
    -
    \mathbf{J}^\top\mathbf{W}F,
    \label{eq_damped_normal_equation}
\end{equation}
where all quantities are evaluated at iteration $k$,
$\mathbf{I}$ is the identity matrix, and $\lambda\geq0$ is the
damping parameter. Once the step is accepted, the state is updated by
\begin{equation}
    \mathbf{x}(k+1)
    =
    \mathbf{x}(k)
    \boxplus
    \boldsymbol{\Delta}(k).
    \label{eq_state_update}
\end{equation}
The optimization then proceeds to the next nonlinear iteration, where the TSDF query structure and the valid RSS sample pairs are recomputed from the updated state.

The complete iterative procedure is summarized in Algorithm~\ref{alg_flowchart}. The surface and space terms constrain endpoint agreement and the internal consistency of the surrounding field, respectively, while the RSS consistency term relates off-surface samples to their measured endpoints through the ray geometry and TSDF-derived surface estimates. Throughout optimization, the sample--endpoint pairings remain fixed, whereas the corresponding surface estimates are recomputed by TSDF projection at each nonlinear iteration. The consistency and proximity checks then select the valid RSS sample pairs for the current linearization, allowing the geometric coupling to adapt as the poses and TSDF values are jointly optimized.

For efficient implementation, TSDF queries during optimization are accelerated using a sparse hash-block structure, enabling fast evaluation of interpolated field values and gradients. After optimization, the updated TSDF is stored in an OpenVDB-based sparse volumetric representation \citep{museth2013openvdb} for memory-efficient downstream mapping and large-scale SLAM.

\section{SDF-SLAM}

Directly applying SDF-BA to a long sequence would result in a
high-dimensional nonlinear problem with strong coupling between a large number of pose and map variables, increasing both computational cost and sensitivity to initialization. To enable scalable and robust optimization in large-scale environments, we propose SDF-SLAM, an SDF-based submap joining framework that combines local joint pose--map optimization with global submap optimization.
SDF-SLAM confines the joint pose--map optimization to bounded
submaps, within which SDF-BA jointly optimizes the local poses and TSDF
map state. At the global level, only the coordinate-frame poses of
spatially overlapping submaps are jointly optimized using
cross-submap SDF constraints. This decomposition substantially reduces
the dimensionality and nonlinear coupling of the optimization while
preserving local joint pose--map optimization, thereby improving
scalability and reducing the risk of convergence to poor local
solutions over long trajectories. The complete procedure is
summarized in Algorithm~\ref{alg_sdf_slam}.

The input sequence is first partitioned into partially overlapping
submaps. For each submap $a$, SDF-BA, as defined in
Algorithm~\ref{alg_flowchart}, jointly optimizes the local trajectory
$\hat{\mathbf{x}}_a^P$ and TSDF map state
$\hat{\mathbf{x}}_a^D$. Let $D_a^{\star}(\cdot)$ denote the optimized
local TSDF associated with $\hat{\mathbf{x}}_a^D$. Its zero-level
surface is extracted to construct cross-submap SDF constraints.
Consecutive submaps are connected by transition edges, while
spatially overlapping submaps are coupled through direct SDF
constraints. After global submap optimization, the optimized submap
poses are combined with the locally optimized trajectories to recover
the global trajectory, from which the final global TSDF map is
reconstructed.

At the global level, each submap $a$ is represented by a graph node whose state is its coordinate frame pose $\mathbf{T}_a^{S}\in\mathrm{SE}(3)$. The pose $\mathbf{T}_a^{S}$ transforms a point from the local coordinate frame of submap $a$ to the global coordinate frame. The collection of all submap poses is denoted by
\begin{equation}
\mathcal{T}^{S}
=
\left\{
\mathbf{T}_a^{S}
\right\}_{a=0}^{N_S-1},
\end{equation}
where $N_S$ denotes the total number of submaps. Each submap pose is initialized from the pose of its anchor frame in the input odometry trajectory. This initial pose is denoted by $\mathbf{T}_a^{S,0}$.

The zero-level surface of the optimized local TSDF
$D_a^{\star}(\cdot)$ is represented by
\begin{equation}
    \mathcal{S}_a
    =
    \left\{
        \mathbf{s}_a^m
        \,\middle|\,
        D_a^{\star}\!\left(\mathbf{s}_a^m\right)=0
    \right\}.
    \label{eq:submap_surface_set}
\end{equation}
The extracted surface points are used to construct direct SDF
residuals between spatially overlapping submaps.

Candidate submap pairs are identified according to the spatial
overlap predicted by the initial submap poses. For each submap, an
oriented bounding box is constructed from its local geometry and
transformed into the global frame using $\mathbf{T}_a^{S,0}$.
Pairs with intersecting bounding boxes are retained as overlap
candidates~\citep{ericson2004real}, forming the candidate set
$\mathcal{C}_{\mathrm{overlap}}$.

Each candidate edge $(a,b)$ is further validated against the corresponding local TSDFs before being included in the submap graph.
For a surface point $\mathbf{s}_a^m\in\mathcal{S}_a$, its position
in the coordinate frame of submap $b$ is
\begin{equation}
    \mathbf{q}_{ab}^m
    =
    \left(\mathbf{T}_b^{S}\right)^{-1}
    \mathbf{T}_a^{S}
    \mathbf{s}_a^m.
    \label{eq:submap_point_transformation}
\end{equation}
The corresponding direct SDF residual is
\begin{equation}
    r_{ab}^{m}\!\left(\mathcal{T}^{S}\right)
    =
    D_b^{\star}\!\left(\mathbf{q}_{ab}^m\right).
    \label{eq:direct_sdf_submap_residual}
\end{equation}
For each candidate edge $(a,b)$, $\Omega_{ab}$ contains the indices of
source surface points whose transformed positions lie within the
observed truncation region of submap $b$ and admit valid TSDF
interpolation. Candidate edges with insufficient valid support or excessive SDF inconsistency are rejected, and the accepted edges form
$\mathcal{E}_R$. Since the residual is directional, $(a,b)$ and
$(b,a)$ are validated independently, and either or both edges
may be included in $\mathcal{E}_R$. The point indices in
$\Omega_{ab}$ remain fixed during optimization, while the residuals
and Jacobians are reevaluated as the submap poses are updated.

\begin{algorithm}[t]
\caption{SDF-SLAM Framework}
\label{alg_sdf_slam}
\SetKwInput{KwInput}{Input}
\SetKwInput{KwOutput}{Output}
\SetKwFunction{FPartition}{SubmapPartition}
\SetKwFunction{FLocalBA}{SDF-BA}
\SetKwFunction{FExtract}{ExtractSurface}
\SetKwFunction{FInitPose}{InitAnchorPose}
\SetKwFunction{FOverlap}{DetectOverlap}
\SetKwFunction{FValidate}{BuildSDFEdges}
\SetKwFunction{FOptimize}{OptimizeSubmaps}
\SetKwFunction{FRecover}{RecoverTrajectory}
\SetKwFunction{FGlobalMap}{GlobalTSDF}
\small
\SetAlgoLined
\DontPrintSemicolon

\KwInput{Point clouds $\mathbb{P}$ and odometry $\mathbb{O}$}
\KwOutput{Global poses $\hat{\mathbb{T}}$ and global TSDF $\hat{D}(\cdot)$}

$\{\mathbb{P}_a,\mathbb{O}_a\}_{a=0}^{N_S-1} \leftarrow$ \FPartition{$\mathbb{P},\mathbb{O}$}\;

\tcp{Local submap optimization}
\For{$a\leftarrow 0$ \KwTo $N_S-1$}{
$\hat{\mathbf{x}}_a^P,\hat{\mathbf{x}}_a^D \leftarrow$ \FLocalBA{$\mathbb{P}_a,\mathbb{O}_a$}\;
\tcp{SDF-BA as defined in Algorithm~\ref{alg_flowchart}}
$\mathcal{S}_a \leftarrow$ \FExtract{$D_a^{\star}(\cdot)$}\;
$\mathbf{T}_a^{S,0} \leftarrow$ \FInitPose{$\mathbb{O},a$}\;
$\mathcal{M}_a \leftarrow \{D_a^{\star}(\cdot),\mathcal{S}_a,\mathbf{T}_a^{S,0}\}$\;
}

\tcp{Submap graph construction}
$\mathbb{M} \leftarrow \{\mathcal{M}_a\}_{a=0}^{N_S-1}$\;

$\mathcal{E}_O \leftarrow \{(a,a+1)\mid a=0,\ldots,N_S-2\}$\;

$\{\mathbf{T}_{ab}^{S,O}\}_{(a,b)\in\mathcal{E}_O}
\leftarrow
\{(\mathbf{T}_a^{S,0})^{-1}\mathbf{T}_b^{S,0}\}_{(a,b)\in\mathcal{E}_O}$\;
\tcp{Odometry transition references}

$\mathcal{C}_{\mathrm{overlap}} \leftarrow$ \FOverlap{$\mathbb{M}$}\;

$\mathcal{E}_R,\{\Omega_{ab}\} \leftarrow$ \FValidate{$\mathcal{C}_{\mathrm{overlap}},\mathbb{M}$}\;
\tcp{Direct SDF residuals in \eqref{eq:direct_sdf_submap_residual}}

\tcp{Global submap optimization}
$\mathcal{T}^{S,0} \leftarrow \{\mathbf{T}_a^{S,0}\}_{a=0}^{N_S-1}$\;

$\hat{\mathcal{T}}^S \leftarrow$ \FOptimize{$\mathcal{T}^{S,0},\mathcal{E}_O,\{\mathbf{T}_{ab}^{S,O}\},\mathcal{E}_R,\{\Omega_{ab}\}$}\;
\tcp{Minimize the objective in \eqref{eq:sgo_objective}}

$\hat{\mathbb{T}} \leftarrow$ \FRecover{$\hat{\mathcal{T}}^S,\{\hat{\mathbf{x}}_a^P\}_{a=0}^{N_S-1}$}\;

$\hat{D}(\cdot) \leftarrow$ \FGlobalMap{$\mathbb{P},\hat{\mathbb{T}}$}\;

\textbf{return} $\hat{\mathbb{T}},\hat{D}(\cdot)$\;

\end{algorithm}

The resulting direct SDF term is
\begin{equation}
    f_R^S\!\left(\mathcal{T}^{S}\right)
    =
    \sum_{(a,b)\in\mathcal{E}_R}
    \sum_{m\in\Omega_{ab}}
    \left\|
        r_{ab}^{m}\!\left(\mathcal{T}^{S}\right)
    \right\|^2.
    \label{eq:direct_sdf_sgo}
\end{equation}

In addition to the direct SDF term, consecutive submaps are
connected by odometry-derived transition edges. The
corresponding edge set is
\begin{equation}
    \mathcal{E}_O
    =
    \left\{
        (a,a+1)
        \,\middle|\,
        a=0,\ldots,N_S-2
    \right\}.
\end{equation}
For each $(a,b)\in\mathcal{E}_O$, the relative motion between the
corresponding anchor frames in the input odometry trajectory defines
a fixed reference transform:
\begin{equation}
    \mathbf{T}_{ab}^{S,O}
    =
    \left(\mathbf{T}_a^{S,0}\right)^{-1}
    \mathbf{T}_b^{S,0}.
    \label{eq:submap_transition_measurement}
\end{equation}
During global optimization, this reference remains fixed, while the
current inter-submap transform is
$\left(\mathbf{T}_a^S\right)^{-1}\mathbf{T}_b^S$. The transition
term penalizes their discrepancy:
\begin{equation}
    f_O^S\!\left(\mathcal{T}^{S}\right)
    =
    \sum_{(a,b)\in\mathcal{E}_O}
    \left\|
        \log\!\left(
            \left(\mathbf{T}_{ab}^{S,O}\right)^{-1}
            \left(\mathbf{T}_a^{S}\right)^{-1}
            \mathbf{T}_b^{S}
        \right)
    \right\|^2.
    \label{eq:sgo_odometry_term}
\end{equation}
This term preserves the relative-motion information provided by the
input odometry and complements the direct SDF constraints between
overlapping submaps.

Therefore, the complete submap graph objective is defined as
\begin{equation}
    f_{\mathrm{SGO}}\!\left(\mathcal{T}^{S}\right)
    =
    w_R^S f_R^S\!\left(\mathcal{T}^{S}\right) + w_O^S f_O^S\!\left(\mathcal{T}^{S}\right),
    \label{eq:sgo_objective}
\end{equation}
where $w_R^S$ and $w_O^S$ weight the direct SDF and transition terms,
respectively. The objective is minimized using an iterative nonlinear
least-squares procedure analogous to that employed in SDF-BA, but with
the optimization variables restricted to the submap poses
$\mathcal{T}^{S}$. The first submap pose is fixed to remove the global
gauge freedom.

The local and global optimization stages are coupled through the
optimized submap representations. SDF-BA jointly optimizes the local
trajectory and TSDF map state, and the resulting
$D_a^{\star}(\cdot)$ is used to construct the cross-submap
SDF constraints. The global stage therefore adjusts only the submap
coordinate-frame poses while preserving the locally optimized
geometry within each submap. After optimization, the global pose of
each scan is recovered by composing its locally optimized pose with the
optimized pose of the corresponding submap frame. The final global
TSDF is then reconstructed from the recovered global trajectory and
the input point cloud measurements.

\section{Experimental Results}

In this section, we conduct extensive experiments on public datasets spanning diverse environments to evaluate both the core SDF-BA formulation and the complete SDF-SLAM system. The evaluation covers both short trajectory segments and complete trajectories. It further assesses robustness to degraded odometry inputs and observations, as well as computational efficiency. The proposed method is compared with representative state-of-the-art approaches across these evaluation settings.

\subsection{Compared Methods and Evaluation Metrics}\label{sec::exp_evaluation}
We first introduce the state-of-the-art methods used for comparison
in the different evaluation settings, followed by the evaluation metrics for pose accuracy and map quality.

\textbf{\textit{State-of-the-Art Methods:}} We compare the proposed method with representative state-of-the-art approaches spanning LiDAR-inertial odometry, plane-based BA, dense map-based BA, SDF-based SLAM, and TSDF-based mapping.

FAST-LIO2 \citep{xu2022fast} serves as the LiDAR-inertial odometry baseline. The plane-based BA approaches include BALM2 \citep{liu2023efficient} and HBA \citep{liu2023large}, while the dense map-based BA methods include Occupancy-SLAM \citep{wang2025occupancy} and MAD-BA \citep{cwian2025mad}. Voxgraph \citep{reijgwart2019voxgraph} is included as an SDF-based SLAM method. For map quality evaluation, we further compare against three TSDF mapping approaches: VDBFusion \citep{vizzo2022vdbfusion}, Voxfield \citep{pan2022voxfield}, and nvblox \citep{millane2024nvblox}. Since these TSDF mapping methods perform mapping without pose estimation, they are provided with the same fixed FAST-LIO2 poses for map construction.

Occupancy-SLAM is evaluated only on short trajectory segments because
its 3D formulation incurs substantial memory consumption, making it
impractical for most of the large-scale sequences considered in this
work. Conversely, HBA and Voxgraph are included only in the
complete-trajectory evaluation, while VDBFusion, Voxfield, and nvblox
are used exclusively for map-quality evaluation.

\textbf{\textit{Pose Accuracy Evaluation:}} Pose accuracy is evaluated using the Root Mean Square Error (RMSE) of
the absolute pose error (APE). The estimated trajectories are aligned
with the ground truth trajectories using EVO~\citep{grupp2017evo},
following the evaluation protocols adopted in
\cite{rosinol2021kimera} and \cite{liu2023large}.

\textbf{\textit{Quantitative Map Quality Evaluation:}}
Following common reconstruction evaluation
protocols~\citep{pan2024pin,zhong2023shine,zhang2025curl}, we
evaluate reconstruction quality using mean accuracy (M. Acc.),
mean completeness (M. Comp.), the bidirectional Chamfer-$L_1$
distance (C-L1), and the F-score at a distance threshold of
$20\,\mathrm{cm}$ ($F_{20}$). For all methods except Voxgraph,
maps are reconstructed at a spatial resolution of $5\,\mathrm{cm}$.
Voxgraph is evaluated using the native TSDF map produced by its
online SLAM pipeline. The reference map is sampled at
$5\,\mathrm{cm}$ for metric computation. For each method and
sequence, the estimated trajectory is aligned with the temporally
associated ground truth trajectory using a rigid transformation in
$\mathrm{SE}(3)$. This alignment is composed with the fixed
transformation between the ground truth trajectory and reference map
frames to place the reconstructed map in the reference map frame.
No method-specific map-level ICP alignment is performed, so that trajectory-induced mapping errors are preserved in the evaluation and the reported metrics reflect the overall SLAM performance.

\begin{table}[t]
\centering
\small
\setlength{\tabcolsep}{1 pt}
\caption{Characteristics of the evaluated datasets and sequences.}
\label{tab:dataset_characteristics}
\begin{tabular}{@{}llcccc@{}}
\toprule
Dataset & Sequence & \#Poses & Len. (m) & Environment & Platform \\
\midrule

\multirow{3}{*}{NC 20}
& Short Experiment & 15302 & 1609 & campus/park & \multirow{3}{*}{handheld} \\
& Quad Dynamic     & 3987  & 479  & campus      & \\
& Parkland Mount   & 5009  & 696  & campus/park & \\
\midrule

\multirow{7}{*}{NC 21}
& Quad Hard        & 1880  & 235  & campus      & \multirow{7}{*}{handheld} \\
& Stairs           & 1190  & 57   & indoor      & \\
& Cloister         & 2788  & 429  & cloister    & \\
& Park             & 15722 & 2396 & campus/park & \\
& Maths Hard       & 2440  & 321  & campus      & \\
& \makecell[l]{Underground\\Hard}
                    & 1908  & 237  & indoor      & \\
\midrule

Arche
& -- & 3615 & 339 & \makecell{disaster\\area} & UAV \\
\midrule

\multirow{4}{*}{BG}
& 1005\_05 & 6142 & 567
& \multirow{4}{*}{garden}
& \multirow{4}{*}{\makecell{wheeled\\robot}} \\
& 1006\_03 & 5947 & 686 & & \\
& 1008\_03 & 6150 & 750 & & \\
& 1018\_13 & 1938 & 201 & & \\
\midrule

\multirow{4}{*}{\makecell{MARS-\\LVIG}}
& AMtown 01    & 13543 & 4918 & town    & \multirow{4}{*}{UAV} \\
& AMvalley 01  & 12010 & 4380 & valley  & \\
& HKairport 01 & 7798  & 2056 & airport & \\
& HKisland 01  & 7510  & 1873 & island  & \\
\bottomrule
\end{tabular}

\vspace{1mm}
\caption*{\footnotesize
\#Poses: number of poses; Len.: trajectory length.}
\end{table}

\subsection{Datasets}

We evaluate the proposed method on five public datasets covering
diverse environments, sensing platforms, and LiDAR configurations:
the Newer College 2020 dataset (NC 20)~\citep{ramezani2020newer},
the Newer College 2021 dataset (NC 21)~\citep{zhang2021multicamera},
the Arche dataset~\citep{reijgwart2019voxgraph}, the Botanic Garden
dataset (BG)~\citep{liu2023botanicgarden}, and the MARS-LVIG
dataset~\citep{li2024mars}. Their main characteristics and the
evaluated sequences are summarized in Table~\ref{tab:dataset_characteristics}.

The datasets are briefly described as follows:
\begin{itemize}

\item \textbf{NC 20 and NC 21:} These two datasets were collected at New College, Oxford, using handheld Ouster LiDAR sensors, with NC 20 using an OS1-64 and NC 21 using an OS0-128. NC 20 mainly covers outdoor campus and garden environments, while NC 21 contains a broader range of indoor and outdoor scenes, such as buildings, corridors, parks, and underground areas. Survey-grade laser-scanned
reference maps are provided for geometric evaluation.

\item \textbf{Arche:} This dataset was introduced with Voxgraph \citep{reijgwart2019voxgraph} and was collected using an Ouster OS1-64 LiDAR mounted on a hexacopter MAV in a disaster-area environment. It contains a single UAV sequence with ground truth positions provided by RTK-GNSS.

\item \textbf{BG:} The BG dataset was collected in a garden
environment using an all-terrain wheeled robot equipped with a Livox
Avia LiDAR. It contains challenging unstructured outdoor scenes with
vegetation, trails, bridges, grasslands, and riverside areas.
Ground truth trajectories are provided based on a survey-grade 3D
reference map.

\item \textbf{MARS-LVIG:} This large-scale aerial dataset was
collected using a UAV equipped with a Livox Avia LiDAR. It covers
diverse outdoor environments, including an airfield, an island, a
rural town, and a valley. The high flight altitude, long trajectories,
and substantial viewpoint variation make it particularly challenging
for LiDAR-based SLAM. Ground truth positions are provided by RTK, and
reference maps are also available for the evaluated environments.

\end{itemize}

\subsection{Setup}\label{sec::exp_setup}

The compared methods operate at different pose estimation
rates. LiDAR odometry and SLAM systems typically estimate poses
at 10--20 Hz to maintain high-rate state updates. In contrast,
BA-based methods optimize accumulated measurements over a set of
selected poses or keyframes and can therefore be evaluated at a lower
temporal resolution. In our experiments, poses are selected for optimization at 2~Hz for BALM2, MAD-BA, Occupancy-SLAM, HBA, and the proposed method. This setting provides a practical balance between optimization cost and pose estimation accuracy and is consistent with the keyframe-based configurations commonly adopted by these BA methods. For evaluation, all estimated trajectories are temporally associated with the
ground truth trajectories and sampled at the same evaluation frequency.

\begin{table*}[ht]
\small
\centering
\setlength{\tabcolsep}{6.8 pt}
\caption{Quantitative results of translation APE (RMSE, in meters) evaluated on real-world datasets over short sequences.}
\label{tab:pose_short_sequence_comparison_rmse}
\begin{tabular}{llccccccc}
\toprule
Dataset &  Sequence
& FAST-LIO2
& BALM2
& MAD-BA
& Occ.-SLAM
& \makecell{Ours\\(Pose Only)}
& \makecell{Ours\\(w/o RSS)}
& \makecell{Ours\\(Full)}  \\
\midrule

\multirow{4}{*}{NC 20}
& Short Experiment & 0.040 & \textbf{0.025} & \underline{0.026} & 0.033 &0.040  &0.054  &\underline{0.026}  \\
& Quad Dynamic     & 0.046 & \underline{0.034} & 0.045 & 0.046 &0.046  &0.092  &\textbf{0.027}  \\
& Parkland Mount   & 0.097 & 0.090 & 0.092 & \textbf{0.074} &0.097  &\underline{0.081}  &\underline{0.081}  \\
\cmidrule(lr){2-9}
& Average & 0.061 & \underline{0.050} & 0.054 & 0.051 & 0.061 & 0.076 & \textbf{0.045} \\
\midrule

\multirow{7}{*}{NC 21}
& Quad Hard        & 0.038 & \underline{0.030} & 0.032 & 0.031 & 0.038 & 0.038 & \textbf{0.027} \\
& Stairs           & 0.071 & \textbf{0.044} & 0.059 & 0.061 & 0.056 & 0.056 & \underline{0.047} \\

& Cloister & 0.058 & \underline{0.054} & 0.078 & 0.056 & 0.058 & 0.060 & \textbf{0.044} \\

& Park             & 0.036 & \underline{0.028} & 0.030 & 0.032 & 0.036 & 0.031 & \textbf{0.024} \\
& Maths Hard       & 0.041 & 0.037 & 0.039 & \underline{0.035} & 0.037 & 0.038 & \textbf{0.028} \\
& Underground Hard & 0.054 & \textbf{0.052} & \underline{0.053} & 0.054 & 0.054 & \underline{0.053} & \underline{0.053} \\
\cmidrule(lr){2-9}

& Average          & 0.050 & \underline{0.041} & 0.049 & 0.045 & 0.047 & 0.046 & \textbf{0.037} \\

\midrule

\multirow{1}{*}{Arche}
&--             & 0.091 & 0.085 & 0.092 & \underline{0.083} &0.085  &0.091  & \textbf{0.079}  \\
\midrule

\multirow{5}{*}{BG}
& 1005\_05         & 0.198 & 0.065 & 0.068 & \underline{0.062} & 0.198 & 0.198 & \textbf{0.052} \\
& 1006\_03         & 0.052 & \textbf{0.039} & 0.041 & \underline{0.040} & 0.052 & 0.052 & \textbf{0.039} \\
& 1008\_03         & 0.088 & \underline{0.067} & \textbf{0.058} & \textbf{0.058} & 0.088 & 0.088 & \textbf{0.058} \\
& 1018\_13         & 0.098 & 0.060 & 0.071 & \underline{0.050} & 0.098 & 0.098 & \textbf{0.048} \\
\cmidrule(lr){2-9}
& Average          & 0.109 & 0.058 & 0.060 & \underline{0.053} & 0.109 & 0.109 & \textbf{0.049} \\
\midrule

\multirow{5}{*}{MARS-LVIG}
& AMtown 01        & 0.261 & 0.253 & 0.253 & \underline{0.244} &0.261  &0.261  & \textbf{0.227}  \\
& AMvalley 01      & 0.185 & 0.188 & 0.197 & \underline{0.173} &0.185  &0.185  & \textbf{0.161}  \\
& HKairport 01     & 0.130 & \textbf{0.038} & 0.077 & 0.114 &0.130  &0.147  &\underline{0.045}  \\
& HKisland 01      & 0.194 & \underline{0.182} & 0.184 & \textbf{0.181} &0.194  &0.197  & \textbf{0.181}  \\
\cmidrule(lr){2-9}
& Average & 0.193 & \underline{0.165} & 0.178 & 0.178 & 0.193 & 0.198 & \textbf{0.154} \\
\bottomrule
\end{tabular}
\caption*{Occ.-SLAM denotes Occupancy-SLAM. \textbf{Bold} indicates the best performance and \underline{underline} indicates the second best performance.}
\end{table*}

FAST-LIO2 is used to provide the initial odometry for methods that do not estimate their own front-end odometry, including the proposed method, BALM2, MAD-BA, HBA, Occupancy-SLAM, and Voxgraph. The LiDAR scans used by these methods are also motion-compensated using FAST-LIO2, following the setting adopted in \cite{liu2023efficient}. This ensures that differences in the
subsequent optimization are not confounded by different scan
undistortion procedures.

For map quality evaluation, methods that do not natively produce a
dense surface representation are first remapped from the same
motion-compensated LiDAR scans using their respective estimated
trajectories. The resulting point cloud maps are then converted into meshes using the Ball Pivoting Algorithm~\citep{bernardini1999ball}, as adopted in
CURL-SLAM~\citep{zhang2025curl}, to provide a common surface-level
representation for evaluation. This reduces the influence of different native map representations and enables a fairer geometric comparison across methods. For TSDF-based methods, meshes are directly extracted from the zero-level sets of their reconstructed fields.

\subsection{Pose Accuracy Evaluation of SDF-BA}

We first evaluate the proposed core component, SDF-BA, on short
trajectory segments from diverse environments and compare its pose
accuracy with FAST-LIO2, BALM2, MAD-BA, and Occupancy-SLAM. Each
segment consists of the first 500 LiDAR scans at the original scan
rate, with pose selection and evaluation performed at the common
frequency described in Section~\ref{sec::exp_setup}. For \textit{HKairport 01} of MARS-LVIG, ground truth is unavailable over part of the initial trajectory. We therefore select the first 500 consecutive LiDAR scans for which valid ground truth poses are available.

The translation APE results are reported in
Table~\ref{tab:pose_short_sequence_comparison_rmse}. Overall,
\textit{Ours (Full)} achieves the lowest dataset-level APE on all five evaluated datasets and ranks first or second on all 18 individual sequences. This consistency across datasets with different geometric characteristics highlights the ability of SDF-BA to construct observation--map constraints through TSDF queries without requiring explicit primitive extraction or discrete measurement-to-surface correspondence search.

The comparison with BALM2 illustrates the different behavior of
plane-based and field-based optimization. BALM2 remains highly competitive when informative planar structures can be reliably extracted, achieving the lowest APE on \textit{Stairs} and \textit{HKairport 01}. Its relative advantage, however, varies considerably across environments. SDF-BA reduces the APE relative to BALM2 by 14.3\% on \textit{Park} and 14.4\% on \textit{AMvalley 01}, while also achieving a lower average error on every evaluated dataset. This behavior is consistent with the different geometric assumptions of the two formulations: BALM2 constructs constraints from extracted planar primitives, whereas SDF-BA directly evaluates valid TSDF queries and can therefore incorporate local surface geometry that does not need to conform to a predefined primitive class.

MAD-BA provides a more closely related comparison because both methods jointly optimize poses and a dense map state. SDF-BA nevertheless achieves substantially lower errors on several challenging sequences, reducing the APE by 43.6\% on \textit{Cloister} and 41.6\% on \textit{HKairport 01}. A key distinction lies in how observations interact with the map. MAD-BA relies on explicit measurement-to-surfel associations, whereas SDF-BA evaluates observations directly against the TSDF, where valid queries provide metric geometric information throughout the truncation region. The consistently lower dataset-level errors support the effectiveness of this field-based formulation as an alternative to optimization based on discrete measurement-to-surfel correspondences.

Occupancy-SLAM offers another complementary comparison because it also jointly optimizes robot poses and voxel-wise variables. It remains competitive on several individual sequences, particularly in less structured environments, but SDF-BA achieves the lowest dataset-level error on all five datasets. Unlike the previous comparison, the principal distinction here is the geometric quantity being optimized: Occupancy-SLAM represents occupancy evidence, whereas SDF-BA optimizes a metric SDF. The latter directly encodes the magnitude and direction of local geometric misalignment around observed surfaces. The results favor the proposed formulation across the evaluated datasets, while the computational implications of their map states are examined below.

To directly assess the central components of SDF-BA, we further evaluate two variants. \textit{Ours (w/o RSS)} removes the RSS consistency term while retaining the remaining formulation. \textit{Ours (Pose Only)} optimizes only the robot poses during each optimization stage and subsequently updates the TSDF using the optimized
poses, rather than jointly treating the poses and TSDF values as variables within the same NLLS problem. Directly applying the parameters of \textit{Ours (Full)} to this pose-only formulation leads to substantially degraded performance. We therefore tune its parameters separately to obtain its best performance and treat it as a pose-only baseline rather than a strict ablation.

The comparison between \textit{Ours (Pose Only)} and \textit{Ours (Full)} provides direct evidence for the contribution of joint pose--map optimization. \textit{Ours (Pose Only)} fails to improve upon the FAST-LIO2 initialization on 15 of the 18 evaluated sequences, while \textit{Ours (Full)} achieves a lower APE than
\textit{Ours (Pose Only)} on all 18 sequences. In the pose-only
variant, pose estimation and TSDF updating are performed sequentially,
so consistency between the current trajectory and TSDF values is not
jointly enforced within the same objective. This distinction is
particularly relevant to the RSS consistency term, whose residuals couple the ray--surface geometry with signed distance values away from the zero-level surface.
Jointly adjusting the poses and TSDF values allows these quantities to
adapt under the same objective, providing a more consistent geometric
state for the RSS consistency term.

The RSS ablation further evaluates the role of directly constraining the SDF beyond the zero-level surface. In \textit{Ours (w/o RSS)}, the surface term remains the only geometric observation term that directly couples range measurements with both poses and map variables. The space and regularization terms constrain the internal consistency of the surrounding TSDF, but do not anchor its off-surface values directly to the measured ray geometry. RSS changes this role by turning the surrounding SDF into an observation-constrained geometric quantity that is jointly coupled with the trajectory. The full formulation improves 16 of the 18 sequences and matches the remaining two. The effect is particularly pronounced on BG, where the average APE decreases from 0.109~m to 0.049~m, corresponding to a 55.0\% reduction. At the sequence level, the reductions reach 73.7\% on \textit{1005\_05}, 70.7\% on \textit{Quad Dynamic}, and 69.4\% on \textit{HKairport 01}. These results demonstrate that maintaining a consistent TSDF around the observed surface is not sufficient to obtain the same pose accuracy as directly coupling that field to the range observations. By relating ray--surface geometry to off-surface TSDF values, RSS extends the geometric observation model from the zero-level surface into the surrounding truncation region, allowing this volumetric geometry to directly participate in pose--map optimization.

\newcommand{\spc}{\phantom{$^{*}$}}
\newcommand{\bad}{\rlap{$^{*}$}\spc}

\begin{table*}[ht!]
\centering
\small
\setlength{\tabcolsep}{1.7pt}
\caption{Translation APE (RMSE, in meters) on the NC21 and BG datasets under degraded odometry inputs and observations.}
\label{tab:robustness_nc21_bg}

\begin{tabular}{llccccc|ccccc}
\toprule
\multirow{2}{*}{Dataset} & \multirow{2}{*}{Sequence}
& \multicolumn{5}{c|}{Mild}
& \multicolumn{5}{c}{Severe} \\
\cmidrule(lr){3-7} \cmidrule(lr){8-12}
&
& FAST-LIO2 & BALM2 & MAD-BA & Occ.-SLAM & Ours
& FAST-LIO2 & BALM2 & MAD-BA & Occ.-SLAM & Ours \\
\midrule

\multirow{6}{*}{NC 21}
& Quad Hard
& 0.135\spc
& \underline{0.048}\spc
& 0.055\spc
& 0.094\spc
& \textbf{0.032}\spc
& 1.025\spc
& 1.271\bad
& \underline{0.107}\spc
& 0.310\spc
& \textbf{0.068}\spc \\

& Cloister
& 0.403\spc
& \underline{0.144}\spc
& 0.562\bad
& 0.377\spc
& \textbf{0.102}\spc
& 0.818\spc
& \underline{0.367}\spc
& 0.415\spc
& 0.437\spc
& \textbf{0.266}\spc \\

& Park
& 0.183\spc
& 0.058\spc
& \underline{0.035}\spc
& 0.057\spc
& \textbf{0.028}\spc
& 0.234\spc
& 0.065\spc
& \underline{0.051}\spc
& 0.093\spc
& \textbf{0.036}\spc \\

& Maths Hard
& 0.147\spc
& 0.048\spc
& \underline{0.047}\spc
& 0.061\spc
& \textbf{0.039}\spc
& 1.177\spc
& 0.665\spc
& \underline{0.064}\spc
& 0.146\spc
& \textbf{0.044}\spc \\

& Underground Hard
& 0.180\spc
& 0.093\spc
& \textbf{0.056}\spc
& 0.078\spc
& \underline{0.067}\spc
& 1.064\spc
& 0.952\spc
& \underline{0.126}\spc
& 0.180\spc
& \textbf{0.096}\spc \\

\cmidrule(lr){2-12}
& Average
& 0.210\spc
& \underline{0.078}\spc
& 0.151\spc
& 0.133\spc
& \textbf{0.054}\spc
& 0.864\spc
& 0.664\spc
& \underline{0.153}\spc
& 0.233\spc
& \textbf{0.102}\spc \\

\midrule

\multirow{5}{*}{BG}
& 1005\_05
& 0.467\spc
& 0.167\spc
& \underline{0.103}\spc
& 0.140\spc
& \textbf{0.078}\spc
& 1.261\spc
& 1.104\spc
& \underline{0.213}\spc
& 1.911\bad
& \textbf{0.203}\spc \\

& 1006\_03
& 0.159\spc
& 0.073\spc
& \underline{0.063}\spc
& 0.125\spc
& \textbf{0.057}\spc
& 0.374\spc
& 0.383\bad
& \underline{0.110}\spc
& 0.272\spc
& \textbf{0.088}\spc \\

& 1008\_03
& 0.087\spc
& \underline{0.079}\spc
& 0.123\bad
& 0.136\bad
& \textbf{0.065}\spc
& 0.330\spc
& $\times$\spc
& 0.352\bad
& \underline{0.142}\spc
& \textbf{0.121}\spc \\

& 1018\_13
& 0.218\spc
& 0.097\spc
& 1.415\bad
& \underline{0.081}\spc
& \textbf{0.078}\spc
& 0.665\spc
& 0.347\spc
& \underline{0.256}\spc
& 0.324\spc
& \textbf{0.209}\spc \\

\cmidrule(lr){2-12}
& Average
& 0.233\spc
& \underline{0.104}\spc
& 0.426\spc
& 0.121\spc
& \textbf{0.070}\spc
& 0.658\spc
& $\times$\spc
& \underline{0.233}\spc
& 0.662\spc
& \textbf{0.155}\spc \\

\bottomrule
\end{tabular}

\caption*{Occ.-SLAM denotes Occupancy-SLAM. $^{*}$ indicates performance degradation relative to the corresponding FAST-LIO2 initialization, and $\times$ indicates that the method failed to complete the optimization. \textbf{Bold} and \underline{underline} indicate the best and second-best results, respectively.}
\end{table*}

Overall, the short-segment experiments provide direct support for the central contributions of SDF-BA. Our formulation achieves consistently high accuracy across diverse environments without requiring explicit primitive extraction or discrete measurement-to-surface correspondence search. In addition, the controlled comparisons further show that joint pose--map optimization and RSS provide
complementary benefits: joint optimization maintains consistency
between the trajectory and the optimized distance field, while RSS consistency
allows the observations to constrain TSDF values beyond the
zero-level surface. Their combination yields the most consistent
accuracy across the evaluated sequences.

\subsection{Robustness Assessment}

Although Table~\ref{tab:pose_short_sequence_comparison_rmse} evaluates the compared methods and our SDF-BA across diverse environments, it does not directly assess their robustness to degraded odometry inputs and observations. This is particularly relevant for BA-based methods, which rely on the front end to provide both the initial trajectory and the point cloud measurements used in subsequent optimization.

Under the nominal setting, FAST-LIO2 operates with the original
inertial and LiDAR measurements and provides reliable odometry and
motion-compensated point clouds. In practical deployments, however,
IMU and LiDAR measurements can be degraded by sensor noise, bias
drift, platform vibration, and adverse sensing conditions, affecting
both the initialization and the geometric observations supplied to
the backend. We therefore conduct additional experiments on NC 21 and
BG by perturbing both the raw IMU measurements and LiDAR ranges. These
datasets represent two distinct geometric conditions: NC 21 contains
relatively structured indoor and outdoor environments, whereas BG is
dominated by less structured natural scenes with vegetation, trails,
and weaker geometric regularity.

Specifically, zero-mean Gaussian noise and slowly varying bias drift are added to the gyroscope and accelerometer measurements, while zero-mean Gaussian noise is added to the range of each LiDAR return before reconstructing the corresponding 3D point, with its original beam direction unchanged. Two degradation levels are considered. In the mild setting, the standard deviations of the gyroscope, accelerometer, and LiDAR range noise are set to $0.08$~rad/s, $0.50$~m/s$^2$, and $0.10$~m, respectively. In the severe setting, they are increased to $0.12$~rad/s, $0.90$~m/s$^2$, and $0.15$~m, together with stronger IMU bias drift. These perturbations degrade the odometry inputs and observations while retaining usable inputs for the subsequent BA-based optimization. The \textit{Stairs} sequence is excluded because, under these perturbations, FAST-LIO2 can no longer provide a reliable initialization. All other experimental settings
remain unchanged from those used in Table~\ref{tab:pose_short_sequence_comparison_rmse}.

The resulting translation APE values are reported in
Table~\ref{tab:robustness_nc21_bg}. The proposed method exhibits the
most consistent performance across both datasets and degradation
levels. Under the mild setting, it achieves the lowest APE on four
of the five NC 21 sequences and ranks second on
\textit{Underground Hard}, yielding the best average APE of
$0.054$~m. Under severe degradation, it achieves the best result on
all five NC 21 sequences, with an average APE of $0.102$~m compared
with $0.153$~m for the second-best method, MAD-BA. The difference is
particularly pronounced on \textit{Quad Hard}, \textit{Maths Hard},
and \textit{Underground Hard}: FAST-LIO2 degrades to $1.025$~m,
$1.177$~m, and $1.064$~m, respectively, whereas SDF-BA reduces these
errors to $0.068$~m, $0.044$~m, and $0.096$~m.

A similar trend is observed on BG. SDF-BA achieves the best result on
all four sequences under both mild and severe degradation, with average
APE values of $0.070$~m and $0.155$~m, outperforming the second-best
averages by 32.7\% and 33.5\%, respectively. In contrast, several
competing methods deteriorate relative to the degraded FAST-LIO2
initialization: for example, MAD-BA increases from $0.218$~m to
$1.415$~m on \textit{1018\_13} under mild degradation, while
Occupancy-SLAM increases from $1.261$~m to $1.911$~m on
\textit{1005\_05} under severe degradation, and BALM2 fails on
\textit{1008\_03}. SDF-BA, in contrast, consistently improves the
initialization and achieves the lowest error in every BG case,
demonstrating stronger robustness in unstructured natural environments.

The behavior of the competing methods further illustrates the
difference in robustness. BALM2 remains competitive under mild
degradation, particularly on NC 21, but becomes considerably less
stable as the perturbation increases. On \textit{Quad Hard}, for
example, its APE increases to $1.271$~m under the severe setting,
exceeding the corresponding FAST-LIO2 initialization of $1.025$~m.
Its degradation is more evident on BG, where it fails to complete
the optimization on \textit{1008\_03} under the severe setting.
This behavior is consistent with the stronger dependence of
plane-based optimization on sufficiently reliable planar structure
and associations.

\begin{table*}[t]
\centering
\small
\setlength{\tabcolsep}{2pt}
\renewcommand{\arraystretch}{1.05}
\caption{Translation APE (RMSE, in meters) on complete real-world sequences.}
\label{tab:complete_trajectory_ape}
\begin{tabular*}{\textwidth}{@{\extracolsep{\fill}}llccccccc@{}}
\toprule
Dataset & Sequence & FAST-LIO2 & BALM2 & MAD-BA & HBA & Voxgraph & \makecell{Ours\\(w/o SDF-BA)} & \makecell{Ours\\(Full)} \\
\midrule

\multirow{4}{*}{NC 20}
& Short Experiment & 0.413 & $\times$ & 0.425 & 0.334 & 0.635 & \underline{0.285} & \textbf{0.265} \\
& Quad Dynamic & 0.120 & \underline{0.114} & \underline{0.114} & 0.121 & 0.197 & 0.120 & \textbf{0.110} \\
& Parkland Mount & 0.199 & 0.151 & \underline{0.145} & 0.158 & 0.251 & 0.199 & \textbf{0.143} \\
\cmidrule(lr){2-9}
& Average & 0.244 & $\times$ & 0.228 & 0.204 & 0.361 & \underline{0.201} & \textbf{0.173} \\
\midrule

\multirow{7}{*}{NC 21}
& Quad Hard & 0.049 & 0.046 & 0.071 & 0.055 & \underline{0.041} & 0.049 & \textbf{0.038} \\
& Stairs & 0.077 & \textbf{0.057} & 0.087 & 0.412 & 0.073 & 0.067 & \underline{0.058} \\
& Cloister & 0.273 & 2.111 & \underline{0.108} & 0.283 & 0.255 & 0.273 & \textbf{0.099} \\
& Park & 0.294 & $\times$ & 0.276 & \underline{0.255} & 0.294 & 0.294 & \textbf{0.165} \\
& Maths Hard & 0.059 & 0.057 & \underline{0.054} & 0.057 & 0.060 & 0.059 & \textbf{0.050} \\
& Underground Hard & \underline{0.052} & \underline{0.052} & \underline{0.052} & 0.072 & 0.053 & \underline{0.052} & \textbf{0.051} \\
\cmidrule(lr){2-9}
& Average & 0.134 & $\times$ & \underline{0.108} & 0.189 & 0.129 & 0.132 & \textbf{0.077} \\
\midrule

Arche
& -- & 0.526 & \underline{0.485} & 0.502 & 0.494 & 0.523 & 0.525 & \textbf{0.459} \\
\midrule

\multirow{5}{*}{BG$^{*}$}
& 1005\_05 & 0.450 & 0.706 & \textbf{0.159} & 0.367 & 0.448 & 0.218 & \underline{0.214} \\
& 1006\_03 & 2.500 & $\times$ & 2.270 & 2.326 & 2.505 & \underline{2.226} & \textbf{2.198} \\
& 1008\_03 & 1.613 & $\times$ & 1.324 & 1.524 & 1.477 & \underline{0.852} & \textbf{0.373} \\
& 1018\_13 & 0.315 & 0.185 & 0.178 & 0.204 & 0.312 & \underline{0.146} & \textbf{0.143} \\
\cmidrule(lr){2-9}
& Average & 1.220 & $\times$ & 0.983 & 1.105 & 1.186 & \underline{0.861} & \textbf{0.732} \\
\midrule

\multirow{5}{*}{MARS-LVIG$^{*}$}
& AMtown 01 & 1.373 & $\times$ & \underline{0.917} & 1.179 & 1.048 & 1.066 & \textbf{0.888} \\
& AMvalley 01 & 4.817 & $\times$ & 4.234 & 6.827 & 6.041 & \underline{1.113} & \textbf{1.045} \\
& HKairport 01 & 0.632 & $\times$ & \underline{0.158} & 0.380 & 0.416 & 0.632 & \textbf{0.132} \\
& HKisland 01 & 0.532 & $\times$ & \underline{0.363} & 0.380 & 0.403 & 0.411 & \textbf{0.338} \\
\cmidrule(lr){2-9}
& Average & 1.839 & $\times$ & 1.418 & 2.192 & 1.977 & \underline{0.806} & \textbf{0.601} \\

\bottomrule
\end{tabular*}
\vspace{1mm}
\begin{minipage}{0.98\textwidth}
\footnotesize
$\times$ indicates that the method failed to complete the optimization.
$^{*}$ marks datasets for which all optimization-based methods use the loop closure corrected odometry initialization.
\textbf{Bold} and \underline{underlining} indicate the best and second-best results, respectively.
\end{minipage}
\end{table*}

MAD-BA generally exhibits stronger robustness than BALM2 and is the
second-best method on average under both severe settings. Nevertheless,
its behavior remains less uniform across sequences. For example, it
degrades relative to the corresponding initialization on
\textit{Cloister} under mild perturbation, and on
\textit{1008\_03} and \textit{1018\_13} of BG. These results are
consistent with the sensitivity of explicit surfel associations to
perturbations in the local surface geometry. Occupancy-SLAM also
remains effective on several degraded sequences, but its corrections
are less consistent across the two datasets. In the severe
\textit{1005\_05} case, for example, its APE increases from the
FAST-LIO2 initialization of $1.261$~m to $1.911$~m.

Taken together, these experiments demonstrate the robustness of
SDF-BA to degraded odometry inputs and observations across both NC21 and BG. Unlike plane- and surfel-based formulations, SDF-BA does not rely on extracting planar features or establishing discrete measurement-to-surface correspondences, whose reliability can be affected by perturbations in the input geometry. Instead, surface and ray-based observations jointly constrain the poses and TSDF values through the distance information and local geometry encoded in the field. This reduces dependence on primitive extraction and discrete correspondence selection while preserving geometric constraints around observed surfaces, consistent with the more uniform pose corrections observed across the two datasets.

\begin{figure}[t]
\centering
\includegraphics[width=\linewidth]{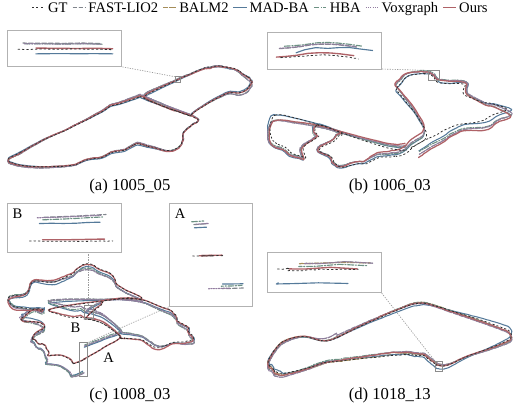}
\caption[Trajectory comparison on BG.]
{Complete-trajectory comparison on the BG dataset. Enlarged regions
highlight representative local trajectory deviations. Ours denotes
the full SDF-SLAM system.}
\label{fig_BG_trajectories}
\end{figure}

\subsection{Complete-Trajectory Pose Accuracy Evaluation of SDF-SLAM}

The short-segment evaluation in
Table~\ref{tab:pose_short_sequence_comparison_rmse} assesses the
core SDF-BA formulation independently of large-scale submap joining.
We next evaluate the complete SDF-SLAM framework on 18 real-world
sequences spanning diverse environments, trajectory lengths, and
spatial extents. All optimization-based methods use the same
FAST-LIO2 trajectory as their common odometry input. For BG and
MARS-LVIG, where long-term odometry drift is more pronounced, loop
candidates are detected using the BTC descriptor~\citep{yuan2024btc},
and the FAST-LIO2 trajectories are corrected through loop-closure
optimization implemented in Ceres. The corrected trajectories are
used both as the common input to all optimization-based methods and
as the FAST-LIO2 baseline reported in the comparison.

Depending on the sequence scale, SDF-SLAM operates with either a
single map or multiple submaps. For a single map sequence, the
framework reduces to SDF-BA over the complete trajectory. Otherwise, the trajectory is partitioned into multiple submaps, within which the poses and TSDF states are jointly optimized by SDF-BA before global submap optimization. We additionally evaluate
\textit{Ours (w/o SDF-BA)}, which removes this local joint
optimization while retaining the same submap partitioning and global optimization. In this variant, each local TSDF is constructed by fusing the corresponding scans using the input odometry poses. For single map sequences, the variant therefore reduces to the input odometry.

\begin{figure}[t]
\centering
\includegraphics[width=\linewidth]{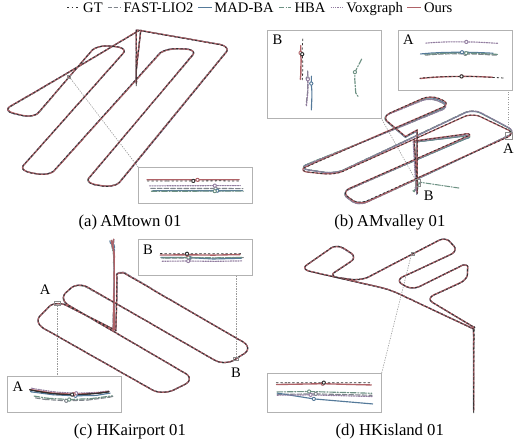}
\caption[Trajectory comparison on MARS-LVIG.]
{Complete-trajectory comparison on the MARS-LVIG dataset. Enlarged
regions highlight representative local trajectory deviations. Ours
denotes the full SDF-SLAM system.}
\label{fig_MARSD_trajectories}
\end{figure}

Table~\ref{tab:complete_trajectory_ape} reports the translation APE.
\textit{Ours (Full)} achieves the lowest dataset-level error on all
five datasets and the lowest APE on 16 of the 18 individual
sequences. It ranks second only on \textit{Stairs} and
\textit{1005\_05}, where BALM2 and MAD-BA achieve the lowest errors,
respectively. On \textit{Short Experiment}, the best competing result
is 0.334~m from HBA, whereas SDF-SLAM achieves 0.265~m. On
\textit{Park}, the corresponding error decreases from 0.255~m for
HBA to 0.165~m. These correspond to reductions of 20.7\% and
35.3\%, respectively. Larger improvements are observed on several
long trajectories in BG and MARS-LVIG.

On BG, SDF-SLAM achieves the lowest APE on three of the four
sequences and the best dataset-level accuracy. The lowest average
APE among the other state-of-the-art methods is 0.983 m, achieved
by MAD-BA, compared with 0.732 m for SDF-SLAM, corresponding to a
25.5\% reduction. The difference is particularly pronounced on
\textit{1008\_03}: SDF-SLAM reduces the APE from the best competing
result of 1.324 m to 0.373 m, a reduction of 71.8\%.
Figure~\ref{fig_BG_trajectories}(c) shows that several competing
trajectories progressively deviate from the ground truth through the
highlighted regions, while the proposed trajectory remains closely
aligned with the reference. Moreover, the APE of
\textit{Ours (w/o SDF-BA)} is 0.852 m, which is further reduced to
0.373 m by the full system, demonstrating the substantial contribution
of local joint pose--map optimization on this challenging sequence.

The advantage becomes even more evident on the large-scale aerial
sequences of MARS-LVIG, where SDF-SLAM achieves the lowest APE on
all four sequences. Its average error is 0.601 m, compared with
1.418 m for MAD-BA, the best-performing method among the other
state-of-the-art approaches, yielding a 57.6\% reduction. A
particularly challenging case is \textit{AMvalley 01}. As shown in
Figure~\ref{fig_MARSD_trajectories}(b), several competing trajectories
develop large, coherent offsets from the ground truth over extended
parts of the trajectory, whereas SDF-SLAM remains closely aligned
with the reference in the highlighted regions. This substantial
difference is also reflected quantitatively: the best competing APE
is 4.234 m, while SDF-SLAM achieves 1.045 m, reducing the error by
75.3\%. The result demonstrates the ability of SDF-SLAM to constrain
large-scale trajectory inconsistency that persists over long spatial
extents.

The comparison between \textit{Ours (w/o SDF-BA)} and
\textit{Ours (Full)} directly evaluates the contribution of local
joint pose--map optimization within SDF-SLAM. Enabling SDF-BA reduces
the APE on all 18 sequences. On NC21, the average error decreases
from 0.132 m to 0.077 m. Particularly large improvements are observed
on \textit{Cloister}, where the error decreases from 0.273 m to
0.099 m, on BG \textit{1008\_03}, from 0.852 m to 0.373 m, and on
MARS-LVIG \textit{HKairport 01}, from 0.632 m to 0.132 m. The latter
corresponds to a 79.1\% reduction. These results show that optimizing
the local poses together with the TSDF state is not only effective
on the short trajectory segments evaluated previously, but remains
important after being incorporated into the complete submap-based
system. By optimizing the internal pose--map geometry before global
submap optimization, SDF-BA provides more geometrically consistent
submaps for the subsequent global stage.

Voxgraph is the most closely related submap-based method because it
also optimizes SDF constraints between overlapping TSDF
submaps~\citep{reijgwart2019voxgraph}. The comparison therefore
isolates an important difference in how the submaps themselves are
constructed and optimized. In SDF-SLAM, the poses and TSDF state
within each submap are jointly optimized before global submap
optimization, so internal pose--map inconsistency can be reduced
before the submap coordinate frames are aligned globally. Voxgraph,
by comparison, operates on fixed submap geometry during global
optimization. The consistently lower errors of the full SDF-SLAM system, together with the improvement from
\textit{Ours (w/o SDF-BA)} to \textit{Ours (Full)}, demonstrate the
benefit of optimizing the local pose--map geometry before submap joining.

MAD-BA provides one of the strongest comparisons because it also
jointly optimizes poses and map geometry. It achieves the lowest APE on \textit{1005\_05} and remains competitive on several other
sequences, but its errors increase substantially on challenging cases such as \textit{1008\_03} and \textit{AMvalley 01}. Unlike MAD-BA, which uses explicit measurement-to-surfel
correspondences, the proposed method evaluates observations directly against the interpolated TSDF. This avoids discrete surface correspondence search and allows valid observations to contribute wherever TSDF support is available within the truncation region.

\begin{table*}[t]
\centering
\small
\setlength{\tabcolsep}{2.2pt}
\renewcommand{\arraystretch}{1.05}
\caption{Quantitative map reconstruction results on the NC20 and NC21 datasets.}
\label{tab:nc_map_evaluation}

\begin{tabular*}{\textwidth}{
@{\extracolsep{\fill}}
ll
cccc
cccccc@{}}
\toprule
\multirow{3}{*}{Metric}
& \multirow{3}{*}{Method}
& \multicolumn{4}{c}{NC20}
& \multicolumn{6}{c}{NC21} \\
\cmidrule(lr){3-6}
\cmidrule(lr){7-12}
&
& \multicolumn{3}{c}{Sequence}
& \multirow{2}{*}{\textbf{Avg.}}
& \multicolumn{5}{c}{Sequence}
& \multirow{2}{*}{\textbf{Avg.}} \\
\cmidrule(lr){3-5}
\cmidrule(lr){7-11}
&
& \shortstack{Short\\Experiment}
& \shortstack{Quad\\Dynamic}
& \shortstack{Parkland\\Mount}
&
& \shortstack{Quad\\Hard}
& \shortstack{Maths\\Hard}
& Cloister
& Stairs
& Park
& \\
\midrule

\multirow{9}{*}{M. Acc. (cm) $\downarrow$}
& FAST-LIO2
& 13.98 & 7.79 & 11.89 & 11.22
& 4.27 & 7.89 & 15.45 & 6.58 & 12.49 & 9.34 \\
& BALM2
& -- & 7.63 & 10.95 & --
& 4.01 & 7.80 & 14.21 & 7.35 & -- & -- \\
& MAD-BA
& 14.42 & 7.14 & 10.93 & 10.83
& 4.64 & 7.71 & \underline{8.08} & 7.28 & 12.35 & 8.01 \\
& HBA
& \underline{12.11} & 8.59 & 10.72 & 10.47
& 4.46 & 7.85 & 15.49 & 12.90 & 12.16 & 10.57 \\

\addlinespace[1pt]
& Voxgraph
& 16.51 & 9.26 & 14.58 & 13.45
& 5.85 & 9.63 & 12.07 & 7.75 & 14.73 & 10.01 \\
& VDBFusion
& 12.24 & \underline{6.47} & 11.37 & \underline{10.03}
& \textbf{3.30} & \underline{6.70} & 12.03 & 6.06 & 11.55 & 7.93 \\
& Voxfield
& 13.61 & 6.75 & \textbf{9.97} & 10.11
& \underline{3.46} & \textbf{6.50} & 12.16 & 5.46 & \underline{11.35} & \underline{7.79} \\
& nvblox
& 14.14 & 8.11 & 11.09 & 11.11
& 4.18 & 7.06 & 13.37 & \underline{5.00} & 11.52 & 8.23 \\
& Ours
& \textbf{9.40} & \textbf{6.40} & \underline{10.40} & \textbf{8.73}
& 3.72 & 7.38 & \textbf{7.68} & \textbf{4.56} & \textbf{8.60} & \textbf{6.39} \\

\midrule

\multirow{9}{*}{M. Comp. (cm) $\downarrow$}
& FAST-LIO2
& 14.55 & 8.26 & 22.98 & 15.26
& 7.04 & 7.52 & 21.66 & 12.31 & 13.43 & 12.39 \\
& BALM2
& -- & 8.31 & 22.00 & --
& 6.71 & 7.52 & 27.39 & 14.00 & -- & -- \\
& MAD-BA
& 16.17 & \underline{8.00} & 22.13 & 15.43
& 7.38 & 7.37 & \underline{13.72} & 13.94 & 13.31 & \underline{11.14} \\
& HBA
& \underline{12.71} & 8.92 & 21.75 & \underline{14.46}
& 7.20 & 7.57 & 22.59 & 22.65 & \underline{12.93} & 14.59 \\

\addlinespace[1pt]
& Voxgraph
& 25.81 & 9.17 & 20.38 & 18.45
& 6.54 & 9.88 & 16.77 & 13.17 & 16.56 & 12.58 \\
& VDBFusion
& 18.29 & 11.62 & 37.41 & 22.44
& 11.06 & 10.69 & 25.83 & 22.35 & 15.36 & 17.06 \\
& Voxfield
& 23.74 & 11.23 & 32.76 & 22.58
& 10.53 & 10.79 & 29.66 & 15.63 & 16.47 & 16.62 \\
& nvblox
& 19.17 & 8.14 & \underline{17.57} & 14.96
& \underline{5.74} & \underline{7.13} & 25.80 & \underline{10.17} & 13.89 & 12.55 \\
& Ours
& \textbf{10.86} & \textbf{6.73} & \textbf{15.86} & \textbf{11.15}
& \textbf{5.15} & \textbf{5.92} & \textbf{13.35} & \textbf{8.12} & \textbf{8.08} & \textbf{8.12} \\

\midrule

\multirow{9}{*}{C-L1 (cm) $\downarrow$}
& FAST-LIO2
& 14.26 & 8.02 & 17.44 & 13.24
& 5.66 & 7.71 & 18.55 & 9.44 & 12.96 & 10.86 \\
& BALM2
& -- & 7.97 & 16.47 & --
& 5.36 & 7.66 & 20.80 & 10.68 & -- & -- \\
& MAD-BA
& 15.30 & \underline{7.57} & 16.53 & 13.13
& 6.01 & 7.54 & \underline{10.90} & 10.61 & 12.83 & \underline{9.58} \\
& HBA
& \underline{12.41} & 8.75 & 16.24 & \underline{12.47}
& 5.83 & 7.71 & 19.04 & 17.77 & \underline{12.55} & 12.58 \\

\addlinespace[1pt]
& Voxgraph
& 21.16 & 9.21 & 17.48 & 15.95
& 6.20 & 9.76 & 14.42 & 10.46 & 15.64 & 11.30 \\
& VDBFusion
& 15.27 & 9.04 & 24.39 & 16.23
& 7.18 & 8.70 & 18.93 & 14.21 & 13.46 & 12.50 \\
& Voxfield
& 18.68 & 8.99 & 21.37 & 16.35
& 7.00 & 8.64 & 20.91 & 10.55 & 13.91 & 12.20 \\
& nvblox
& 16.66 & 8.13 & \underline{14.33} & 13.04
& \underline{4.96} & \underline{7.09} & 19.59 & \underline{7.59} & 12.70 & 10.39 \\
& Ours
& \textbf{10.13} & \textbf{6.57} & \textbf{13.13} & \textbf{9.94}
& \textbf{4.43} & \textbf{6.65} & \textbf{10.52} & \textbf{6.34} & \textbf{8.34} & \textbf{7.26} \\

\midrule

\multirow{9}{*}{$F_{20}$ (\%) $\uparrow$}
& FAST-LIO2
& 76.96 & 94.87 & 77.10 & 82.98
& 97.15 & 93.22 & 68.91 & \underline{90.83} & 79.43 & 85.91 \\
& BALM2
& -- & 95.03 & 79.18 & --
& 97.16 & 93.09 & 69.54 & 86.37 & -- & -- \\
& MAD-BA
& 73.70 & 95.32 & 79.40 & 82.81
& 96.06 & 93.33 & \textbf{87.38} & 86.65 & 80.34 & \underline{88.75} \\
& HBA
& \underline{84.03} & 94.76 & \underline{79.73} & \underline{86.17}
& 97.14 & 93.01 & 69.26 & 75.84 & \underline{81.45} & 83.34 \\

\addlinespace[1pt]
& Voxgraph
& 57.98 & 91.46 & 68.19 & 72.54
& 95.68 & 86.11 & 74.01 & 86.92 & 69.06 & 82.36 \\
& VDBFusion
& 76.79 & 94.36 & 71.22 & 80.79
& 94.90 & 91.73 & 73.63 & 86.26 & 78.01 & 84.91 \\
& Voxfield
& 67.96 & 94.40 & 73.10 & 78.49
& 95.37 & 91.52 & 65.86 & 89.14 & 76.24 & 83.63 \\
& nvblox
& 69.70 & \underline{95.89} & 79.28 & 81.62
& \underline{97.73} & \underline{93.40} & 61.77 & \textbf{92.14} & 78.45 & 84.70 \\
& Ours
& \textbf{88.29} & \textbf{96.11} & \textbf{81.06} & \textbf{88.49}
& \textbf{97.96} & \textbf{93.85} & \underline{86.20} & \textbf{92.14} & \textbf{93.80} & \textbf{92.79} \\

\bottomrule
\end{tabular*}
\vspace{1mm}
\begin{minipage}{\textwidth}
\textbf{Bold} and \underline{underlining} indicate the best and second-best values.
\end{minipage}
\end{table*}

BALM2 and HBA exhibit a different limitation associated with their
dependence on planar structure. When sufficiently informative planes
are available, their compact geometric representation can provide
competitive estimates. BALM2, for example, achieves the lowest APE
on \textit{Stairs}. However, BALM2 fails to complete eight of the
18 trajectories, including all four MARS-LVIG sequences. In our
experiments, these failures are associated with either insufficient
reliable planar features or the increased memory demand introduced
by finer spatial partitioning required to obtain usable planar
constraints. HBA improves scalability through hierarchical
optimization, but its accuracy still degrades on several sequences
with less informative planar geometry. These results, together with
the robustness evaluation in Table~\ref{tab:robustness_nc21_bg},
show that the effectiveness of plane-based formulations depends
strongly on the availability and reliability of suitable planar
structures. SDF-SLAM avoids this requirement by constructing its
constraints directly from valid TSDF information and can therefore
use the same geometric formulation across both structured and
unstructured environments.

Overall, the complete-trajectory evaluation shows that the accuracy
benefits of SDF-BA are retained within the complete SDF-SLAM
framework. The full system achieves the lowest dataset-level APE on
all five datasets and the lowest sequence-level APE on 16 of the
18 complete trajectories, while the trajectory visualizations further
show that these improvements correspond to reduced local deviations
along challenging large-scale trajectories. These results demonstrate
the consistent trajectory accuracy of SDF-SLAM across substantially
different scene structures and spatial scales.

\subsection{Map Quality Comparison}
In this section, we compare the map reconstruction quality of the
proposed method with representative state-of-the-art SLAM, odometry,
and mapping approaches. In addition to pose estimation methods, we
include three standalone TSDF mapping methods, VDBFusion, Voxfield,
and nvblox, using FAST-LIO2 odometry as their common input. These comparisons assess whether the proposed formulation provides benefits beyond the conventional pipeline of combining a strong odometry front end with a separate TSDF
mapping backend.

For quantitative comparison, we evaluate three NC20 sequences and
five NC21 sequences for which suitable reference maps are available.
All methods follow the common surface-level evaluation protocol
described in Section~\ref{sec::exp_evaluation}. For MAD-BA, we do not directly evaluate its native sparse surfel map, since its limited surface coverage would confound map sparsity with geometric reconstruction quality in metrics such as completeness and Chamfer distance. Instead, the optimized poses are used to remap the original motion-compensated LiDAR scans, and the resulting point cloud is reconstructed using the same Ball Pivoting procedure. This produces a more complete and higher-quality reconstruction for MAD-BA than directly evaluating its native sparse surfel map.

\begin{figure*}[tp]
\centering
\includegraphics[width=0.99\textwidth]{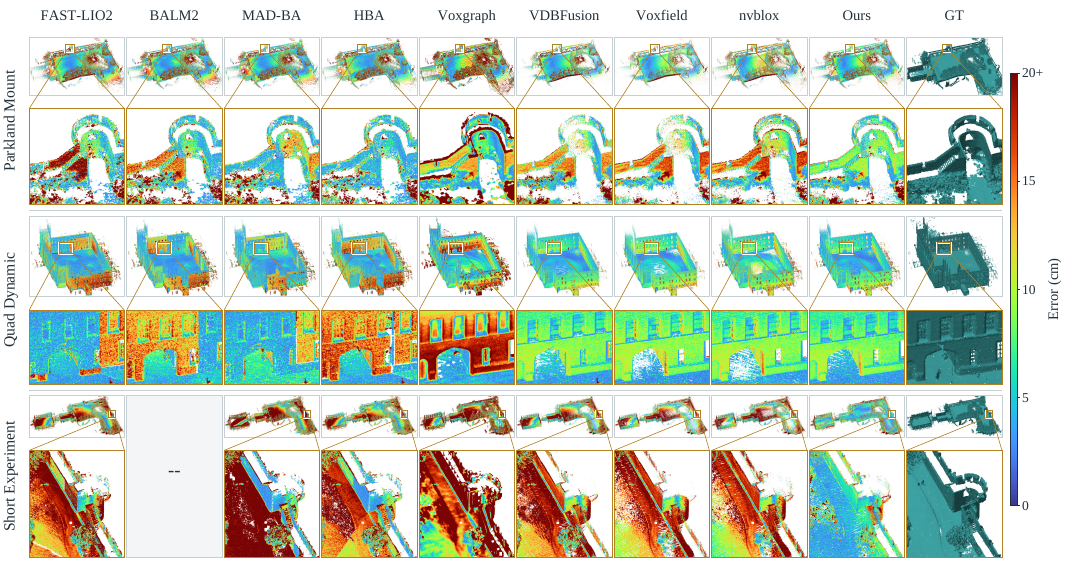}
\caption[Qualitative map reconstruction results on NC20.]
{Qualitative comparison of reconstructed maps on the NC20 dataset.
Colors indicate the distances from the reconstructed surfaces to the ground truth (GT) point cloud. BALM2 fails to produce a valid
reconstruction for the \textit{Short Experiment} sequence.}
\label{fig:nc20_map_comparison}
\end{figure*}

\begin{figure*}[tp]
\centering
\includegraphics[width=0.99\textwidth]{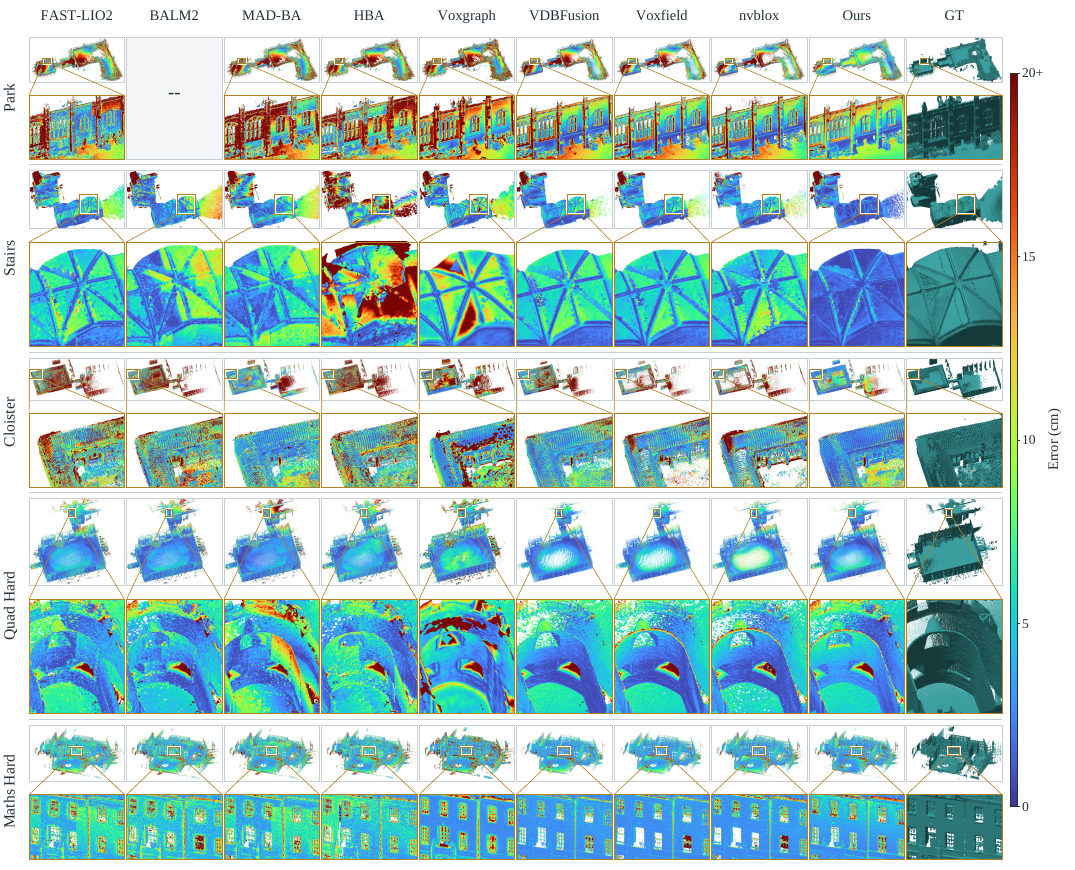}
\caption[Qualitative map reconstruction results on NC21.]
{Qualitative comparison of reconstructed maps on the NC21 dataset.
Colors indicate the distances from the reconstructed surfaces to the ground truth (GT) point cloud. BALM2 fails to produce a valid
reconstruction for the \textit{Park} sequence.}
\label{fig:nc21_map_comparison}
\end{figure*}

Table~\ref{tab:nc_map_evaluation} reports the quantitative results,
while Figures~\ref{fig:nc20_map_comparison} and
\ref{fig:nc21_map_comparison} show the reconstruction results and corresponding spatial error distributions. Our method achieves the best average performance
across all four metrics on both datasets. On NC20, our method achieves the best result in 11 of the 12 comparisons across the three sequences and four metrics, and ranks second in M. Acc. on \textit{Parkland Mount}. On NC21, it achieves the best result in 17 of the 20 comparisons across the five sequences and four metrics, including the lowest M. Comp. and C-L1 on all five sequences. The average C-L1 decreases from 12.47~cm for HBA to 9.94~cm on NC20 and from 9.58~cm for MAD-BA to 7.26~cm on NC21, corresponding to reductions of 20.3\% and
24.2\% relative to the best baseline averages, respectively. BALM2 does not provide a valid reconstruction for \textit{Short Experiment} on NC20 or for \textit{Park} on NC21 because it fails to complete the optimization on these sequences.

The results with the plane-based methods illustrate how their dependence on extracted planar structures affects the final reconstruction across different scene geometries. On NC21, our method reduces the average C-L1 from 12.58~cm for
HBA to 7.26~cm. Even on \textit{Maths Hard}, where HBA remains
competitive, C-L1 decreases from 7.71~cm to 6.65~cm and M. Comp.
from 7.57~cm to 5.92~cm. The difference becomes considerably
larger on \textit{Stairs}: M. Acc., M. Comp., and C-L1 decrease
from 12.90~cm, 22.65~cm, and 17.77~cm with HBA to 4.56~cm,
8.12~cm, and 6.34~cm with our method, respectively, while
$F_{20}$ increases from 75.84\% to 92.14\%. The enlarged view in Figure~\ref{fig:nc21_map_comparison} shows the same trend, with extensive high error regions in the HBA reconstruction and substantially lower errors over the vaulted structure and adjoining surfaces in our result. Together with the BALM2 failures on two of the evaluated sequences, these results are consistent with the different geometric assumptions of the two formulations: plane-based methods depend on the extraction and association of sufficiently informative planar structures, whereas SDF-BA constructs constraints through TSDF queries and can incorporate more general local surface geometry.

MAD-BA is a strong baseline because it jointly optimizes poses and
surfel geometry. Its results on \textit{Cloister} are close to ours: C-L1 is
10.90~cm compared with 10.52~cm, and MAD-BA achieves the higher
$F_{20}$. The enlarged roof and adjoining wall in Figure~\ref{fig:nc21_map_comparison} show low-error regions in both
reconstructions, while residual errors remain near the openings
and on the adjacent ground. The advantage of our method is more pronounced on \textit{Park}, which combines built structures with irregular surrounding geometry. Relative to MAD-BA, M. Acc. decreases from 12.35~cm to 8.60~cm and M. Comp. from 13.31~cm to 8.08~cm. The enlarged building facade in Figure~\ref{fig:nc21_map_comparison} shows less extensive high-error regions on the wall surfaces in our reconstruction than in MAD-BA. The reductions in both metrics indicate closer agreement between the reconstructed and reference surfaces. This sequence-dependent behavior is consistent with the benefit of constraining a shared distance field through surface and ray-derived observations instead of explicit measurement-to-surfel assignments.

\begin{figure*}[tp]
\centering
\includegraphics[width=0.99\textwidth]{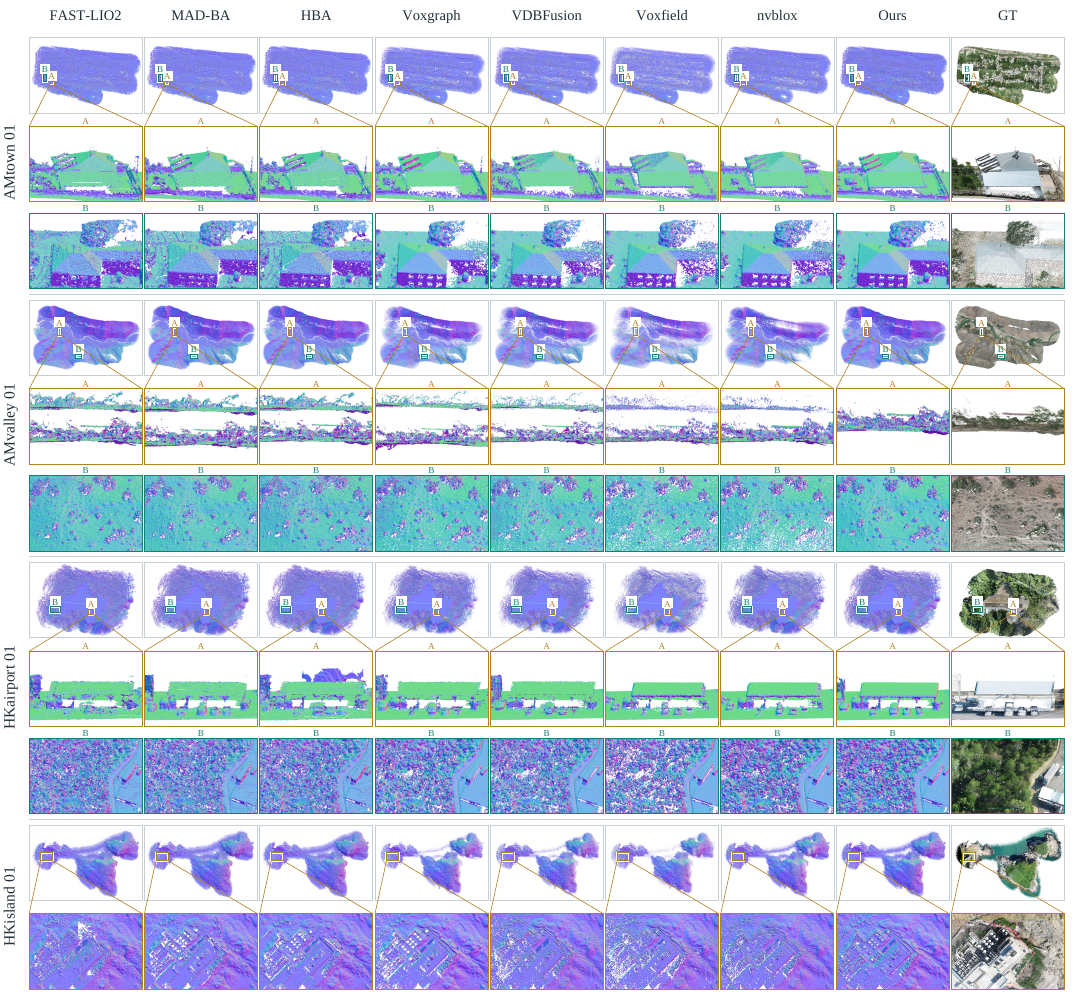}
\caption[Qualitative map reconstruction results on MARS-LVIG.]
{Qualitative comparison on four MARS-LVIG sequences. Overview maps and
enlarged views of the highlighted regions are shown, with the provided
reference maps in the rightmost column.}
\label{fig:mars_map_comparison}
\end{figure*}

The comparison with Voxgraph separates the use of an SDF
representation from the additional local optimization performed
by SDF-BA. On \textit{Parkland Mount}, Voxgraph achieves a C-L1 of
17.48~cm, compared with 13.13~cm for our method. The enlarged view in
Figure~\ref{fig:nc20_map_comparison} shows pronounced high-error
bands along the curved structure and adjoining wall in Voxgraph,
which are substantially reduced in our result. On \textit{Short Experiment}, the difference is larger, with C-L1 decreasing from 21.16~cm to 10.13~cm. The corresponding enlarged view shows substantially lower errors along the wall and across the adjacent ground in our reconstruction than in Voxgraph. Voxgraph aligns SDF submaps through their coordinate frames, whereas SDF-SLAM additionally optimizes the poses and TSDF states within each submap before global
alignment. This distinction is relevant because a rigid
transformation of a submap cannot, by itself, correct its
internal geometric distortion. The observed error reductions
are consistent with the role of local joint optimization rather
than an advantage attributable solely to the use of a TSDF.

The fixed-trajectory reconstruction baselines provide complementary
evidence. In the \textit{Short Experiment} row of
Figure~\ref{fig:nc20_map_comparison}, FAST-LIO2, VDBFusion,
Voxfield, and nvblox retain similar extended high-error regions
on the enlarged wall and ground surfaces despite differences
in local surface appearance. The three standalone mappers achieve C-L1 values between 15.27~cm and 18.68~cm, compared with 10.13~cm for our method. The persistence of a common spatial error pattern across different mappers is consistent with the influence of their shared input trajectory. Changing
the fusion backend can alter the reconstructed surface, but
does not revise the supplied poses. In contrast, the proposed
framework also adjusts the trajectory through joint pose--TSDF
optimization, and the resulting reconstruction substantially
reduces these extended high error regions.

On \textit{Cloister}, Voxfield and nvblox reduce M. Acc. relative
to FAST-LIO2 from 15.45~cm to 12.16~cm and 13.37~cm,
respectively, but increase M. Comp. from 21.66~cm to 29.66~cm
and 25.80~cm. The enlarged view in Figure~\ref{fig:nc21_map_comparison} shows larger gaps in the ground surface adjacent to the building. In contrast, our method achieves the lowest errors in both directions, with an M. Acc. of
7.68~cm and an M. Comp. of 13.35~cm.

We further compare the reconstructed maps on four large-scale
MARS-LVIG sequences in Figure~\ref{fig:mars_map_comparison}. The
provided reference maps are used only for qualitative comparison.
A pronounced difference appears in region A of
\textit{AMvalley 01}, where several baselines exhibit visibly
separated surface layers, while our reconstruction substantially
reduces this duplicated geometry and more closely follows the
reference terrain profile. Region B also shows a more continuous
ground surface than Voxfield and nvblox, whose reconstructions
contain more local gaps.

\begin{table*}[t!]
\centering
\small
\setlength{\tabcolsep}{2.0pt}
\caption{Runtime and peak memory comparison on real-world datasets.}
\label{tab:runtime_memory_long_sequence_comparison}
\begin{tabular}{llc|ccccc|ccccc}
\toprule
\multirow{2}{*}{Dataset}
& \multirow{2}{*}{Sequence}
& \multirow{2}{*}{Duration (s)}
& \multicolumn{5}{c|}{Runtime (s)}
& \multicolumn{5}{c}{Peak Memory (GB)} \\
\cmidrule(lr){4-8}
\cmidrule(lr){9-13}
& &
& BALM2 & MAD-BA & HBA & Voxgraph & Ours
& BALM2 & MAD-BA & HBA & Voxgraph & Ours \\
\midrule

\multirow{3}{*}{NC 20}
& Short Experiment
& 1530
& -- & 7635 & 80 & 1709 & 842$^\dagger$
& -- & 33.26 & \underline{14.24} & 58.42 & \textbf{10.50} \\

& Quad Dynamic
& 398
& 51 & 2031 & 19 & 415 & 444
& 17.63 & 16.04 & \underline{5.28} & \textbf{3.07} & 8.99 \\

& Parkland Mount
& 500
& 67 & 977 & 28 & 602 & 786
& 58.35 & \underline{13.56} & \textbf{6.67} & 24.50 & 18.31 \\

\midrule

\multirow{6}{*}{NC 21}
& Quad Hard
& 187
& 15 & 154 & 4 & 197 & 738
& 37.64 & 4.77 & \textbf{1.81} & \underline{2.68} & 15.27 \\

& Stairs
& 118
& 2 & 49 & 1 & 120 & 182
& 1.99 & 0.77 & \underline{0.49} & \textbf{0.14} & 3.02 \\

& Cloister
& 278
& 18 & 840 & 24 & 303 & 110$^\dagger$
& 35.04 & 11.59 & 5.02 & \underline{4.62} & \textbf{4.13} \\

& Park
& 1571
& -- & 17910 & 59 & 1645 & 606$^\dagger$
& -- & 74.48 & 17.47 & \underline{14.99} & \textbf{11.83} \\

& Maths Hard
& 243
& 18 & 194 & 8 & 256 & 322
& 31.11 & 10.46 & \underline{3.99} & \textbf{3.19} & 17.95 \\

& \mbox{Underground Hard}
& 190
& 10 & 144 & 3 & 196 & 70$^\dagger$
& 28.85 & 4.46 & \underline{1.46} & \textbf{0.81} & 8.86 \\

\midrule

Arche
& --
& 361
& 16 & 2580 & 15 & 373 & 58$^\dagger$
& 39.96 & 9.50 & \underline{4.05} & \textbf{2.50} & 4.59 \\

\midrule

\multirow{4}{*}{BG}
& 1005\_05
& 623
& 143 & 264 & 8 & 671 & 265$^\dagger$
& 76.66 & 3.27 & \underline{2.70} & 55.38 & \textbf{2.09} \\

& 1006\_03
& 577
& -- & 138 & 8 & 624 & 297$^\dagger$
& -- & \underline{2.51} & \textbf{2.37} & 5.42 & 2.61 \\

& 1008\_03
& 625
& -- & 137 & 10 & 672 & 566$^\dagger$
& -- & \textbf{2.49} & \underline{2.55} & 8.36 & 6.72 \\

& 1018\_13
& 179
& 4 & 32 & 3 & 199 & 84$^\dagger$
& 16.92 & \textbf{0.94} & \underline{0.99} & 72.66 & 1.34 \\

\midrule

\multirow{4}{*}{MARS-LVIG}
& AMtown 01
& 1332
& -- & 10511 & 119 & 1830 & 649$^\dagger$
& -- & 32.43 & \underline{22.04} & 72.39 & \textbf{11.44} \\

& AMvalley 01
& 1171
& -- & 3416 & 162 & 1686 & 437$^\dagger$
& -- & 26.86 & \underline{20.43} & 81.79 & \textbf{15.69} \\

& HKairport 01
& 755
& -- & 1926 & 72 & 1092 & 1098
& -- & 21.19 & \textbf{14.97} & 58.88 & \underline{18.92} \\

& HKisland 01
& 637
& -- & 232 & 12 & 720 & 147$^\dagger$
& -- & \textbf{2.38} & \underline{3.23} & 17.33 & 4.37 \\

\bottomrule
\end{tabular}

\vspace{1mm}
\begin{minipage}{0.98\textwidth}
\footnotesize
$\dagger$ indicates that the total runtime of our method is shorter than the data acquisition duration of the corresponding sequence.
For peak memory usage, \textbf{bold} indicates the lowest memory
consumption and \underline{underlining} indicates the second lowest.
-- indicates that the corresponding result is unavailable because the
method failed to complete the optimization.
\end{minipage}
\end{table*}

The building regions provide additional evidence of local geometric
quality. In \textit{AMtown 01}, our reconstruction preserves clear
roof ridges and surface boundaries, while several pose-based
reconstructions exhibit more irregular local geometry. In
\textit{HKairport 01}, the roof outline and surrounding structures
remain well preserved, whereas HBA contains an additional surface
fragment above the roof that is absent from both our result and the
reference map. These observations complement the NC20 and NC21
quantitative results by showing that the proposed method also maintains coherent geometry in larger aerial environments.

Overall, the results demonstrate improved bidirectional
surface agreement on NC20 and NC21, while the MARS-LVIG comparisons
show reduced geometric duplication and coherent local structure in
large-scale aerial scenes.

\subsection{Computational Efficiency and Scalability}

In addition to estimation accuracy, computational efficiency and memory consumption are important considerations for joint pose--map optimization, since introducing map variables can substantially increase the size of the optimization problem. We evaluate the computational characteristics of the proposed approach at two levels. At the system level, the complete SDF-SLAM framework is compared with BALM2, MAD-BA, HBA, and Voxgraph on 18 complete real-world sequences in terms of runtime and peak memory consumption. At the formulation level, SDF-BA is compared with Occupancy-SLAM under controlled variations in the number of input scans and map resolution to examine the scaling behavior of dense joint pose--map optimization. All experiments are conducted on the same workstation equipped with an AMD Ryzen 9 9950X CPU and 96~GB RAM.

Table~\ref{tab:runtime_memory_long_sequence_comparison} reports the runtime and peak memory consumption of the complete systems, together with the acquisition duration of each sequence as a reference for computational throughput. BALM2, MAD-BA, HBA, and SDF-SLAM are evaluated in batch mode. Despite jointly optimizing local poses and explicit TSDF values, SDF-SLAM completes 12 of the 18 sequences in less time than their corresponding acquisition durations, indicating that its average batch-processing throughput exceeds the acquisition rate on these sequences. SDF-SLAM also completes all 18 sequences and achieves the lowest peak memory consumption on six of them, including several long sequences such as \textit{Short Experiment}, NC21 \textit{Park}, \textit{AMtown 01}, and \textit{AMvalley 01}.

The lower runtimes of BALM2 and HBA are consistent with their
plane-based optimization formulations, while HBA additionally
benefits from parallel processing. SDF-SLAM instead jointly
optimizes robot poses and explicit TSDF values representing dense
local geometry, with the map state restricted to active voxels
within each submap. Despite the higher computational cost,
SDF-SLAM remains within the sequence acquisition timescale on
most evaluated trajectories while providing the trajectory and
reconstruction improvements reported in the preceding experiments.

The peak memory results show a different trend. The plane-based
methods do not consistently require less memory than SDF-SLAM:
SDF-SLAM uses less peak memory than BALM2 on nine of the ten
sequences for which BALM2 completes the optimization, and less
than HBA on six of the 18 sequences. The latter include several
long trajectories, with peak memory decreasing from 14.24~GB
to 10.50~GB on \textit{Short Experiment}, from 17.47~GB to
11.83~GB on NC21 \textit{Park}, from 22.04~GB to 11.44~GB
on \textit{AMtown 01}, and from 20.43~GB to 15.69~GB on
\textit{AMvalley 01}. Thus, jointly optimizing poses and
TSDF values for dense local geometry does not necessarily incur
higher system-level peak memory than plane-based optimization.

\begin{figure}[tp]
\centering
\includegraphics[width=\linewidth]{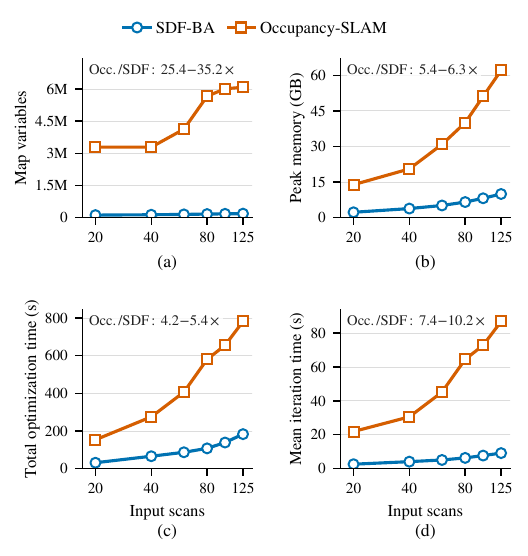}
\caption[Scaling with the number of input scans.]
{Computational scaling of Occupancy-SLAM and SDF-BA on the NC21
\textit{Park} sequence with increasing numbers of input scans.
(a) Number of active map variables. (b) Peak memory consumption.
(c) Total optimization time. (d) Mean optimization time per iteration.
The annotations indicate the range of Occupancy-SLAM to SDF-BA ratios
(Occ./SDF) across the evaluated settings.}
\label{fig:frame_scaling}
\end{figure}

MAD-BA provides a more direct comparison for assessing the computational cost of joint pose--map optimization, as it also optimizes robot poses together with a dense map state. SDF-SLAM is faster on 11 of the 18 sequences, with particularly large differences on several long trajectories. On NC21 \textit{Park}, the runtime decreases from 17910~s for MAD-BA to 606~s for SDF-SLAM, while on \textit{AMtown 01}, it decreases from 10511~s to 649~s, corresponding to speedups of approximately $29.6\times$ and $16.2\times$, respectively. Two design choices contribute to this difference. First, MAD-BA maintains explicit measurement-to-surfel associations that must be updated as the poses and surfel geometry change during optimization. SDF-BA instead evaluates observations directly against the TSDF. For each transformed observation, its location determines the local interpolation support, from which the field values and gradients required for the residuals and Jacobians are evaluated. This avoids repeated discrete measurement-to-surface correspondence searches during iterative optimization. Second, SDF-SLAM confines joint pose--map optimization to individual local submaps, while the global stage optimizes only the submap poses. This bounds the size of the coupled pose--map problems and allows the computational cost of dense joint optimization to remain manageable as the trajectory length increases. Together, these design choices contribute to the favorable runtime behavior of SDF-SLAM relative to MAD-BA on complete trajectories.

Voxgraph provides a complementary system-level comparison because it also represents the environment using TSDF submaps. Its runtime is measured under an online execution protocol and is therefore reported for reference rather than directly compared with the batch runtime of SDF-SLAM. The peak memory results provide a more direct comparison of the two TSDF-based systems. SDF-SLAM requires less peak memory on 12 of the 18 sequences despite additionally performing local joint pose--TSDF optimization. The difference is particularly pronounced on the large-scale MARS-LVIG sequences, where peak memory decreases from 72.39~GB for Voxgraph to 11.44~GB for SDF-SLAM on \textit{AMtown 01}, and from 81.79~GB to 15.69~GB on \textit{AMvalley 01}. After local optimization, SDF-SLAM stores the optimized TSDF submaps using a sparse VDB representation. Together with the submap-based optimization architecture, this allows the complete system to maintain a compact memory footprint while retaining local joint pose--map optimization.

To examine the computational scaling of the joint pose--map
formulation itself, we further compare SDF-BA with
Occupancy-SLAM~\citep{wang2025occupancy}. Both methods optimize robot poses together with dense voxel variables, but differ substantially in the spatial support of their map states. The experiments are conducted on the NC21 \textit{Park} sequence, which contains both structured and irregular geometry. We vary the number of input scans and the voxel resolution and record the number of active map variables, peak memory consumption, total optimization time, and mean optimization time per iteration.

\begin{figure}[tp]
\centering
\includegraphics[width=\linewidth]{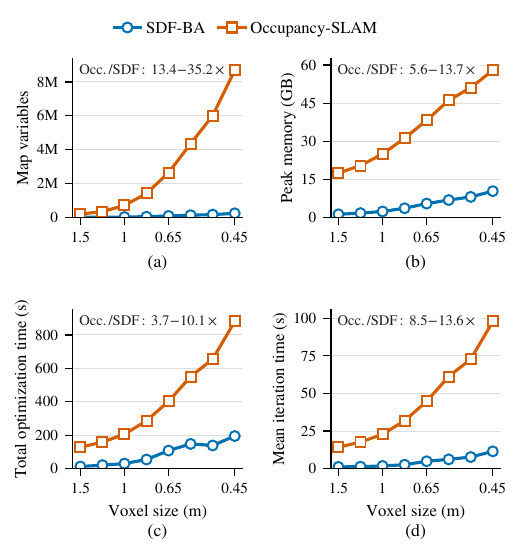}
\caption[Scaling with voxel resolution.]
{Computational scaling of Occupancy-SLAM and SDF-BA on the NC21
\textit{Park} sequence with increasing voxel resolution.
(a) Number of active map variables. (b) Peak memory consumption.
(c) Total optimization time. (d) Mean optimization time per iteration.
The annotations indicate the range of Occupancy-SLAM to SDF-BA ratios
(Occ./SDF) across the evaluated settings.}
\label{fig:resolution_scaling}
\end{figure}

Figure~\ref{fig:frame_scaling} compares the two formulations as the
number of input scans increases. Across the evaluated settings,
Occupancy-SLAM uses 25.4--35.2 times as many active map variables
as SDF-BA, and its peak memory consumption is 5.4--6.3 times that
of SDF-BA. Its total optimization time and mean time per iteration
are 4.2--5.4 and 7.4--10.2 times those of SDF-BA, respectively.
The substantially smaller optimization state of SDF-BA thus
translates into lower memory consumption and computational cost
as more observations are incorporated.

The difference becomes more pronounced as the voxel size decreases,
as shown in Figure~\ref{fig:resolution_scaling}. Occupancy-SLAM uses
13.4--35.2 times as many active map variables and 5.6--13.7 times
more peak memory, while requiring 3.7--10.1 times the total
optimization time and 8.5--13.6 times the mean iteration time.
This scaling difference primarily arises from the map representation:
occupancy-grid optimization maintains variables over a broader mapped
volume, whereas SDF-BA concentrates the optimized TSDF state within
the truncated regions around observed surfaces. The resulting
reduction in optimization variables becomes increasingly important
at finer resolutions as the number of volumetric cells grows rapidly.

This scaling behavior also explains why Occupancy-SLAM is not
included in the complete-trajectory comparison in
Table~\ref{tab:complete_trajectory_ape}. Applying the same voxel
setting to complete trajectories would incur substantially greater
computational and memory costs. The comparison therefore demonstrates
the representation-level advantage of the TSDF formulation for dense
joint pose--map optimization. SDF-SLAM further builds on this
advantage at the system level by confining joint pose--map
optimization to individual submaps, allowing the proposed formulation
to scale from local SDF-BA to complete trajectories.

Overall, the experiments demonstrate the computational advantages of the proposed formulation at both the representation and system levels. At the SDF-BA level, restricting the optimized map state to the truncation regions around observed surfaces substantially reduces the number of active map variables, memory consumption, and
optimization time compared with occupancy grid based joint
pose--map optimization, with the advantage becoming increasingly
pronounced at finer resolutions. At the SDF-SLAM level, localizing dense joint pose--map optimization within individual submaps, together with VDB-based sparse storage of the optimized TSDF submaps, enables this optimization to scale to complete trajectories while maintaining practical runtime and a compact memory footprint. Together with the preceding accuracy results, these experiments show that the proposed formulation achieves an effective balance between dense geometric optimization, computational efficiency, and large-scale scalability.

\section{Conclusion}

In this paper, we presented a geometric formulation for jointly
optimizing robot poses and explicit TSDF values in LiDAR SLAM. Most
geometric SDF-based SLAM systems use the distance field primarily
as a reference for pose estimation, while prior joint optimization
of poses and SDF has mainly been explored through photometric or
appearance-based constraints. In contrast, our formulation establishes
geometric observation--map constraints directly from range
measurements in the SDF domain and jointly optimizes both the
trajectory and the TSDF map. Surface, space, and RSS consistency
terms are integrated into a unified NLLS framework, allowing
both observed surface geometry and off-surface signed distance
information within the relevant TSDF region to contribute to
joint pose--map estimation.

From a broader perspective, this work addresses a key modeling
challenge in SDF-based BA. In feature-, photometric-, surfel-, and
occupancy-based BA formulations, observations can be related to the
corresponding map variables through their native measurement models.
For an SDF, however, range observations provide direct constraints
at the observed surface, while signed distances inferred from local
observations do not, in general, correspond directly to the global
distance field. This local-to-global mismatch makes joint pose--map
optimization in SDF space more difficult to formulate. SDF-BA addresses this challenge by exploiting the intrinsic geometric
properties of SDFs to construct pose-dependent observation--map
constraints, with the RSS consistency term providing the key coupling between range observations, robot poses, and off-surface TSDF values, and RSS consistency validation retaining geometrically valid sample pairs. This enables joint optimization beyond zero-level surface constraints or fixed-map pose estimation.

The proposed formulation also provides a practical approach to dense
LiDAR BA without relying on explicit feature extraction or a
predefined set of geometric primitives. Observations are related to
the map through local TSDF queries, whose interpolated field values
and gradients provide the geometric quantities required for
optimization. Moreover, restricting the optimized TSDF state to the relevant truncation regions yields substantially fewer active map variables than the evaluated occupancy grid formulation. Together with sparse TSDF storage, this enables joint pose--map optimization with practical computational and memory requirements.

To extend the formulation to larger environments, we further
developed SDF-SLAM, a submap-based framework in which SDF-BA jointly
optimizes the poses and TSDF state within each local submap, while
global consistency is established through direct SDF-based
optimization of overlapping submap poses. This decomposition
preserves local joint pose--map optimization while restricting the
global optimization variables to submap poses, enabling the proposed
formulation to scale to longer trajectories and larger environments.

Experiments on multiple public LiDAR datasets demonstrate consistent
improvements in trajectory accuracy and surface reconstruction quality
over representative state-of-the-art methods, while maintaining
practical computational performance. Overall, the results demonstrate that explicit TSDF values can be incorporated directly into geometric BA together with robot poses, providing a framework for coupling dense signed distance representations with multi-view LiDAR range observations.

Future work will investigate more efficient large-scale solvers,
tighter coupling between local and global optimization, and richer
signed distance representations for downstream robotic tasks such as
planning, navigation, and exploration.

\bibliographystyle{SageH}
\bibliography{references}

\appendix

\section{Appendix}

\subsection{Proofs of SDF Geometric Properties}
\label{app:sdf_properties}

The following proofs concern an exact Euclidean SDF $D$, with negative values inside the surface and positive values outside. We assume that the field is differentiable at the query points and that each query has a unique closest point on the zero-level set. Surface-gradient evaluations are restricted to smooth surface points. These assumptions exclude the medial axis and other non-smooth regions of the distance field.

\subsubsection{\textbf{Proof of Lemma~1}}

Let $\mathbf{s}$ be the unique closest surface point to $\mathbf{p}$.
By definition of the signed distance function,
\begin{equation}
    |D(\mathbf{p})|
    =
    \|\mathbf{p}-\mathbf{s}\|.
\end{equation}
At a differentiable point with a unique closest surface point, the
gradient of the SDF points along the corresponding
surface normal. Therefore, for $D(\mathbf{p})\neq 0$,
\begin{equation}
    \nabla D(\mathbf{p})
    =
    \frac{\mathbf{p}-\mathbf{s}}{D(\mathbf{p})}.
\end{equation}
Rearranging gives
\begin{equation}
    \mathbf{s}
    =
    \mathbf{p}
    -
    D(\mathbf{p})\nabla D(\mathbf{p}).
\end{equation}
When $D(\mathbf{p})=0$, $\mathbf{p}$ already lies on the zero-level
surface and $\mathbf{s}=\mathbf{p}$, so the same relation holds.
This proves Lemma~1.

\subsubsection{\textbf{Proof of Lemma~2}}

Let $\mathbf{s}$ be a smooth surface point with outward unit normal $\mathbf{n}_s$. Within a local normal neighborhood in which $\mathbf{s}$ is the unique closest surface point, any point $\mathbf{p}$ in this neighborhood can be written as
\begin{equation}
    \mathbf{p}=\mathbf{s}+d\,\mathbf{n}_s,
    \qquad
    D(\mathbf{p})=d,
\end{equation}
where $d$ is the signed distance to the surface. For $d\neq 0$, Lemma~1 gives
\begin{equation}
    \nabla D(\mathbf{p})
    =
    \frac{\mathbf{p}-\mathbf{s}}{D(\mathbf{p})}
    =
    \mathbf{n}_s.
\end{equation}
The surface normal property gives $\nabla D(\mathbf{s})=\mathbf{n}_s$. Therefore,
\begin{equation}
    \nabla D(\mathbf{p})
    =
    \nabla D(\mathbf{s})
    =
    \mathbf{n}_s.
\end{equation}
The case $d=0$ follows directly from $\mathbf{p}=\mathbf{s}$. This proves Lemma~2.

\subsection{Analytic Jacobian of the RSS Consistency Term}\label{app:jacobians}

This appendix derives the Jacobian of the RSS residual in \eqref{eq_RSS_formal} under the frozen-normal approximation used in the local optimization. The derivation considers a retained observation pair $(\mathbf{p}_{ij},\mathbf{q}_{ijk})$ at frame $i$. Unlike the exact SDF considered in the preceding proofs, $D$ here denotes the trilinearly interpolated TSDF.

For local linearization, the pose increment is written as
\begin{equation}
    \delta\mathbf{x}^P_i =
    \begin{bmatrix}
    \delta\mathbf{t}_i^\top &
    \delta\boldsymbol{\xi}_i^\top
    \end{bmatrix}^\top .
\end{equation}
The translation is updated additively, while the rotation is updated using a left perturbation on $\mathrm{SO}(3)$:
\begin{equation}
    \mathbf{t}_i \leftarrow \mathbf{t}_i+\delta\mathbf{t}_i,
    \qquad
    \mathbf{R}_i \leftarrow \exp(\delta\boldsymbol{\xi}_i^\wedge)\mathbf{R}_i .
\end{equation}
The Jacobian derived below is expressed with respect to this local perturbation.

For the residual associated with $(\mathbf{p}_{ij},\mathbf{q}_{ijk})$, define
\begin{equation}
    \mathbf{q}_{ijk}^\prime
    =
    \mathbf{R}_i\mathbf{q}_{ijk}+\mathbf{t}_i .
\end{equation}
For the retained observation pair, we denote the shared unit normal
introduced in \eqref{eq_shared_normal} by
\begin{equation}
    \mathbf{n}^{C}_{ijk}
    \triangleq
    \mathbf{n}_s.
\end{equation}
For the numerical linearization considered here,
$\widehat{\mathbf{n}}_D(\mathbf{s}_{ijk})$ is used as its
representative estimate. With this notation, the scalar RSS
residual is written as
\begin{equation}
    r^{C}_{ijk}
    =
    -d_{ijk}(\mathbf{n}^{C}_{ijk})^\top
    \mathbf{R}_i\mathbf{n}_{ij}
    -
    D(\mathbf{q}_{ijk}^\prime).
    \label{eq_app_rss_residual}
\end{equation}
Here, \(d_{ijk}\) and \(\mathbf{n}_{ij}\) are fixed quantities computed from the local observation. In each linearization, the retained RSS association and \(\mathbf{n}^{C}_{ijk}\) are held fixed when computing the Jacobian.

\subsubsection{\textbf{Jacobian with Respect to the Pose}}

For compact notation, define
\begin{equation}
    \mathbf{a}_{ij}
    \triangleq
    \mathbf{R}_i\mathbf{n}_{ij},
    \qquad
    \mathbf{b}_{ijk}
    \triangleq
    \mathbf{R}_i\mathbf{q}_{ijk}.
\end{equation}
Then
\begin{equation}
    \mathbf{q}_{ijk}^\prime = \mathbf{b}_{ijk}+\mathbf{t}_i .
\end{equation}
Under the local perturbation above, the first-order variations of \(\mathbf{q}_{ijk}^\prime\) and \(\mathbf{a}_{ij}\) are
\begin{equation}
    \delta\mathbf{q}_{ijk}^\prime
    =
    \delta\mathbf{t}_i
    -
    [\mathbf{b}_{ijk}]_\times\delta\boldsymbol{\xi}_i,
    \label{eq_app_delta_q}
\end{equation}
and
\begin{equation}
    \delta\mathbf{a}_{ij}
    =
    -
    [\mathbf{a}_{ij}]_\times\delta\boldsymbol{\xi}_i,
    \label{eq_app_delta_a}
\end{equation}
where \([\cdot]_\times\) denotes the skew-symmetric matrix.

Using the first-order approximation of the signed distance value at the current linearization point,
\begin{equation}
    \delta D(\mathbf{q}_{ijk}^\prime)
    =
    \nabla D(\mathbf{q}_{ijk}^\prime)^\top
    \delta\mathbf{q}_{ijk}^\prime,
\end{equation}
the variation of the residual in \eqref{eq_app_rss_residual} becomes
\begin{equation}
\begin{aligned}
    \delta r^{C}_{ijk}
    &=
    -d_{ijk}(\mathbf{n}^{C}_{ijk})^\top
    \delta\mathbf{a}_{ij}
    -
    \nabla D(\mathbf{q}_{ijk}^\prime)^\top
    \delta\mathbf{q}_{ijk}^\prime                                      \\
    &=
    d_{ijk}(\mathbf{n}^{C}_{ijk})^\top
    [\mathbf{a}_{ij}]_\times\delta\boldsymbol{\xi}_i
    -
    \nabla D(\mathbf{q}_{ijk}^\prime)^\top\delta\mathbf{t}_i             \\
    &\quad
    +
    \nabla D(\mathbf{q}_{ijk}^\prime)^\top
    [\mathbf{b}_{ijk}]_\times\delta\boldsymbol{\xi}_i .
\end{aligned}
\end{equation}

For compactness, define the rotational Jacobian component as
\begin{equation}
    \mathbf{j}^{\xi}_{ijk}
    \triangleq
    d_{ijk}(\mathbf{n}^{C}_{ijk})^\top[\mathbf{a}_{ij}]_\times
    +
    \nabla D(\mathbf{q}_{ijk}^\prime)^\top[\mathbf{b}_{ijk}]_\times .
\end{equation}
Then
\begin{equation}
    \delta r^{C}_{ijk}
    =
    -\nabla D(\mathbf{q}_{ijk}^\prime)^\top\delta\mathbf{t}_i
    +
    \mathbf{j}^{\xi}_{ijk}\delta\boldsymbol{\xi}_i .
\end{equation}
Since the pose increment is ordered as
\(\delta\mathbf{x}^P_i=[\delta\mathbf{t}_i^\top,\delta\boldsymbol{\xi}_i^\top]^\top\), the Jacobian of one RSS residual with respect to the pose increment is
\begin{equation}
    \frac{\partial r^{C}_{ijk}}
    {\partial \delta\mathbf{x}^P_i}
    =
    \begin{bmatrix}
    -\nabla D(\mathbf{q}_{ijk}^\prime)^\top
    &
    \mathbf{j}^{\xi}_{ijk}
    \end{bmatrix}.
    \label{eq_app_rss_pose_jac}
\end{equation}
The residual \(r^{C}_{ijk}\) depends only on the pose of frame \(i\). Therefore, its Jacobian with respect to all other pose states is zero.

\subsubsection{\textbf{Jacobian with Respect to TSDF Voxel Values}}

During each linearization, the retained RSS association
and $\mathbf{n}^{C}_{ijk}$ are held fixed. Therefore, the RSS residual
depends on the map variables only through the interpolated TSDF value
$D(\mathbf{q}_{ijk}^\prime)$.

Let $\mathcal{V}(\mathbf{q}_{ijk}^\prime)$ denote the voxel indices
involved in the trilinear interpolation at
$\mathbf{q}_{ijk}^\prime$, with interpolation weights
$\alpha_c(\mathbf{q}_{ijk}^\prime)$. Then
\begin{equation}
    D(\mathbf{q}_{ijk}^\prime)
    =
    \sum_{c\in\mathcal{V}(\mathbf{q}_{ijk}^\prime)}
    \alpha_c(\mathbf{q}_{ijk}^\prime)D(\mathbf{m}_c).
    \label{eq_app_tsdf_interp}
\end{equation}

Substituting \eqref{eq_app_tsdf_interp} into
\eqref{eq_app_rss_residual} gives
\begin{equation}
    r^{C}_{ijk}
    =
    -d_{ijk}(\mathbf{n}^{C}_{ijk})^\top\mathbf{a}_{ij}
    -
    \sum_{c\in\mathcal{V}(\mathbf{q}_{ijk}^\prime)}
    \alpha_c(\mathbf{q}_{ijk}^\prime)D(\mathbf{m}_c).
\end{equation}
Therefore, for each active voxel
$c\in\mathcal{V}(\mathbf{q}_{ijk}^\prime)\cap\Omega_D$,
\begin{equation}
    \frac{\partial r^{C}_{ijk}}
    {\partial D(\mathbf{m}_c)}
    =
    -\alpha_c(\mathbf{q}_{ijk}^\prime).
    \label{eq_app_rss_map_jac}
\end{equation}

\end{document}